\documentclass[runningheads]{llncs}

\usepackage{graphicx}
\usepackage{amsmath,amssymb} 
\usepackage{color}

\usepackage{times}
\usepackage{epsfig}
\usepackage{enumerate}
\usepackage[utf8]{inputenc}
\usepackage{amsmath}
\usepackage{dsfont}
\usepackage{array,multirow,graphicx}
\usepackage{booktabs,caption}
\usepackage[table]{xcolor}    
\usepackage{pifont}
\usepackage{hyperref}
\usepackage{multirow,array}
\usepackage{verbatim}
\usepackage{wrapfig,lipsum,booktabs}
\usepackage{rotating}
\usepackage{hyperref}

\usepackage{sidecap} 

\usepackage{makecell}

\usepackage{tikz}
\def\scamera#1#2{
\begin{scope}[shift={#1}, rotate=#2]
    \draw [fill=black](0,0) -- (0.15,0.2) -- (-0.15,0.2) -- cycle;
    \draw [fill=white,ultra thick](0,0) circle (0.1);
\end{scope}
}
\def\fcamera#1#2{
\begin{scope}[shift={#1}, rotate=#2]
    \draw [fill=black](0,0) -- (0.15,0.2) -- (-0.15,0.2) -- cycle;
    \draw [fill=white,ultra thick](0,0) circle (0.1);
\end{scope}
}
\def\bcamera#1#2{
\begin{scope}[shift={#1}, rotate=#2]
    \draw [fill=black](0,0) -- (0.15,0.2) -- (-0.15,0.2) -- cycle;
    \draw [fill=white,ultra thick](0,0) circle (0.1);
\end{scope}
}
\newcommand{\ominicamrotated}{
\begin{tikzpicture}
\fcamera{(0,0)}{45}
\scamera{(0,0)}{135}
\bcamera{(0,0)}{225}
\scamera{(0,0)}{315}
\end{tikzpicture}
}
\newcommand{\ominicam}{
\begin{tikzpicture}
\fcamera{(0,0)}{0}
\scamera{(0,0)}{90}
\bcamera{(0,0)}{180}
\scamera{(0,0)}{270}
\end{tikzpicture}
}

\newcommand{\udacitycam}{
\begin{tikzpicture}
\fcamera{(0,0)}{0}
\scamera{(0,0)}{90}
\scamera{(0,0)}{270}
\end{tikzpicture}
}

\newcommand{\fcam}{
\begin{tikzpicture}
\fcamera{(0,0)}{0}
\end{tikzpicture}
}

\newcommand{\ourdata}{\textbf{Drive360}}
\newcommand{\YES}{\ding{51}}

\definecolor{gred}{RGB}{101,115,40}
\definecolor{red1}{RGB}{255,102,102}
\definecolor{green1}{RGB}{102,178,102}
\definecolor{yellow1}{RGB}{255,128,16}
\definecolor{blue1}{RGB}{48,130,137}
\definecolor{halfred}{RGB}{128,0,0}

\begin{document}
\pagestyle{headings}
\mainmatter
\def\ECCV18SubNumber{933}  

\title{End-to-End Learning of Driving Models with Surround-View Cameras and Route Planners} 

\titlerunning{Driving Models with Surround-View Cameras and Route Planners}
%
\author{Simon Hecker\inst{1} \and
Dengxin Dai\inst{1} \and
Luc Van Gool\inst{1,2}}
%
\authorrunning{S. Hecker, D. Dai and L. Van Gool}
%
\institute{ETH Zurich, Zurich, Switzerland \\
\email{\{heckers,dai,vangool\}@vision.ee.ethz.ch}\\ \and
KU Leuven, Leuven, Belgium}
\maketitle

\begin{abstract}
For human drivers, having rear and side-view mirrors is vital for safe driving. They deliver a more complete view of what is happening around the car. Human drivers also heavily exploit their mental map for navigation. Nonetheless, several methods have been published that learn driving models with only a front-facing camera and without a route planner. This lack of information renders the self-driving task quite intractable. We investigate the problem in a more realistic setting, which consists of a surround-view camera system with eight cameras, a route planner, and a CAN bus reader. In particular, we develop a sensor setup that provides data for a 360-degree view of the area surrounding the vehicle, the driving route to the destination, and low-level driving maneuvers (e.g. steering angle and speed) by human drivers. With such a sensor setup we collect a new driving dataset, covering diverse driving scenarios and varying weather/illumination conditions. Finally, we learn a novel driving model by integrating information from the surround-view cameras and the route planner.  
Two route planners are exploited: 1) by representing the planned routes on OpenStreetMap as a stack of GPS coordinates, and 2) by rendering the planned routes on TomTom Go Mobile and recording the progression into a video. Our experiments show that: 1) 360-degree surround-view cameras help avoid failures made with a single front-view camera, in particular for city driving and intersection scenarios; and 2) route planners help the driving task significantly, especially for steering angle prediction. 
Code, data and more visual results will be made available at \url{http://www.vision.ee.ethz.ch/~heckers/Drive360}.
\end{abstract}

\section{Introduction}
\label{sec:introduction}
Autonomous driving has seen dramatic advances in recent years, for instance for road scene parsing~\cite{Cityscapes,semantic:foggy:scene,BDD100K,daytime:2:nighttime}, lane following \cite{LeCun:driving:05,highway:driving:15,deep:driving}, path planning \cite{perception:path:generation,chen2017brain,paxton2017combining,machines5010006}, and end-to-end driving models~\cite{end:driving:16,end:driving:imitation:18,lidar:end:driving:18,end:driving:eventcamera:18}. By now, autonomous vehicles have driven many thousands of miles and companies aspire to sell such vehicles in a few years. Yet, significant technical obstacles, such as the necessary robustness of driving models to adverse weather/illumination conditions~\cite{semantic:foggy:scene,BDD100K,daytime:2:nighttime} or the capability to anticipate potential risks in advance~\cite{problems:autonomous:vehicle:17,driving:failure:prediction}, must be overcome before assisted driving can be turned into full-fletched automated driving. At the same time, research on the next steps towards `complete' driving systems is becoming less and less accessible to the academic community. We argue that this is mainly due to the lack of large, shared driving datasets delivering more \emph{complete} sensor inputs.   

\begin{figure*}[t]
\includegraphics[width=0.95\linewidth]{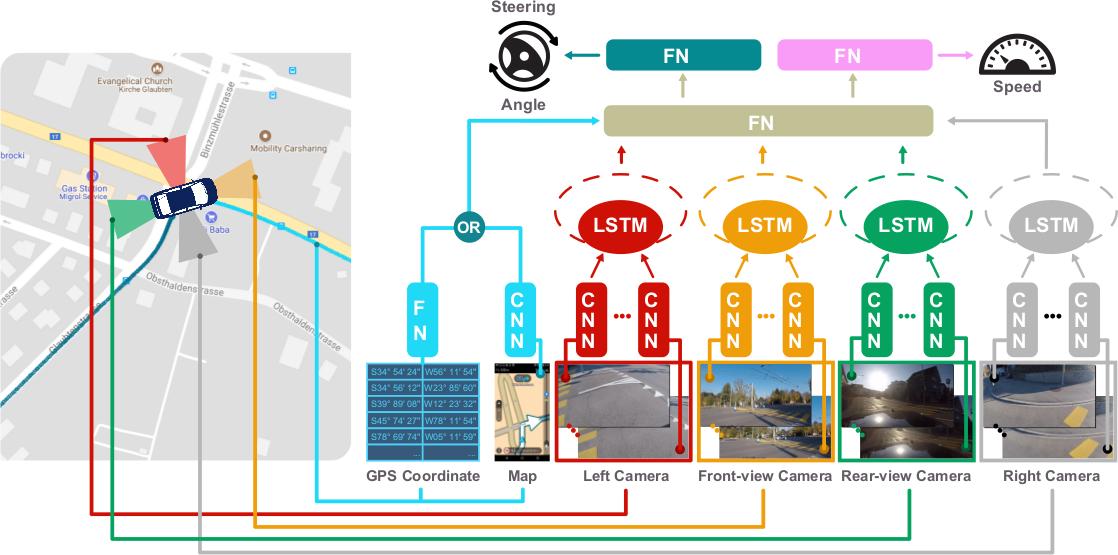}
\caption{An illustration of our driving system. Cameras provide a 360-degree view of the area surrounding the vehicle. The driving maps or GPS coordinates generated by the route planner and the videos from our cameras are synchronized. They are used as inputs to train the driving model. The driving model consists of CNN networks for feature encoding, LSTM networks to integrate the outputs of the CNNs over time; and fully-connected networks (FN) to integrate information from multiple sensors to predict the driving maneuvers.} 
\label{fig:pipeline}
\end{figure*}

\noindent
\textbf{Surround-view cameras and route planners}.
Driving is inarguably a highly visual and intellectual task. Information from all around the vehicle needs to be gathered and integrated to make safe decisions. As a virtual extension to the limited field of view of our eyes, side-view mirrors and a rear-view mirror are used since 1906~\cite{early:mirror} and in the meantime have become obligatory. Human drivers also use their internal maps~\cite{cognitive:maps:rats:men,human:navigation:98} or a digital map to select a route to their destination. Similarly, for automated vehicles, a decision-making system must select a route through the road network from its current position to the requested destination~\cite{AD:Boss:08,AD:systems:algorithm:11,AV:concept:path:12}. 

As said, a single front-view camera is inadequate to learn a safe driving model. It has already been observed in~\cite{network:autonomous:1980} that upon reaching a fork - and without a clearcut idea of where to head for - the model may output multiple widely discrepant travel directions, one for each choice. This would result in unsafe driving decisions, like oscillations in the selected travel direction. Nevertheless, current research often focuses on this setting because it still allows to look into plenty of challenges~\cite{highway:driving:15,nvidia:driving:16,end:driving:16}. This is partly due to the simplicity of training models with a single camera, both in terms of available datasets and the complexity an effective model needs to have. Our work includes a surround-view camera system, a route planner, and a data reader for the vehicle's CAN bus. The setting provides a 360-degree view of the area surrounding the vehicle, a planned driving route, and the `ground-truth' maneuvers by human drivers. Hence, we obtain a learning task similar to that of a human apprentice, where a (cognitive/digital) map gives an overall sense of direction, and the actual steering and speed controls need to be set based on the observation of the local road situation.  

\noindent
\textbf{Driving Models.}
In order to keep the task tractable, we chose to learn the driving model in an end-to-end manner, i.e. to map inputs from our surround-view cameras and the route planner directly to low-level maneuvers of the car. The incorporation of detection and tracking modules for traffic agents (e.g. cars and pedestrians) and traffic control devices (e.g. traffic lights and signs) is future work. 
We designed a specialized deep network architecture which integrates all information from our surround-view cameras and the route planner, and then maps these sensor inputs directly to low-level car maneuvers. See Figure~\ref{fig:pipeline} and the supplemental material for the network's architecture. The route planner is exploited in two ways: 1) by representing planned routes as a stack of GPS coordinates, and 2) by rendering the planned routes on a map and recording the progression as a video.  

Our main contributions are twofold: 1) a new driving dataset of 60 hours, featuring videos from eight surround-view cameras, two forms of data representation for a route planner, low-level driving maneuvers, and GPS-IMU data of the vehicle's odometry; 2) a learning algorithm to integrate information from the surround-view cameras and planned routes to predict future driving maneuvers. Our experiments show that: a) 360-degree views help avoid failures made with a single front-view camera; and b) a route planner also improves the driving significantly. 

\section{Related Work} 
Our work is relevant for 1) driving models, 2) assistive features for vehicles with surround view cameras, 3) navigation and maps, and 4) driving scene understanding.   

\label{sec:related}
\subsection{Driving Models for Automated Cars}
Significant progress has been made in autonomous driving, especially due to the deployment of deep neural networks. Driving models can be clustered into two groups~\cite{deep:driving}: mediated perception approaches and end-to-end mapping approaches, with some  exceptions like~\cite{deep:driving}. Mediated perception approaches require the recognition of all driving-relevant objects, such as lanes, traffic signs, traffic lights, cars, pedestrians, etc.~\cite{kitti:dataset,Cityscapes,3d:object:detection:AD}. Excellent work~\cite{3d:traffic:scene} has been done to integrate such results. This kind of systems developed by the automotive industry represent the current state-of-the-art for autonomous driving. Most use diverse sensors, such as cameras, laser scanners, radar, and GPS and high-definition maps~\cite{autonomous:vehicle:guide:policymakers}. End-to-end mapping methods construct a direct mapping from the sensory input to the maneuvers. The idea can be traced back to the 1980s, when a neural network was used to learn a direct mapping from images to steering angles~\cite{network:autonomous:1980}. Other end-to-end examples are~\cite{LeCun:driving:05,nvidia:driving:16,end:driving:16,lidar:end:driving:18,end:driving:eventcamera:18}. In \cite{end:driving:16}, the authors trained a neural network to map camera inputs directly to the vehicle's ego-motion. Methods have also been developed to explain how the end-to-end networks work for the driving task~\cite{explaining:end:driving:17} and to predict when they fail~\cite{driving:failure:prediction}. Most end-to-end work has been demonstrated with a front-facing camera only. To the best of our knowledge, we present the first end-to-end method that exploits more realistic input. Please note that our data can also be used for mediated perception approaches. 
Recently, reinforcement learning for driving has received increasing attention~\cite{mnih2015humanlevel,reinforcement:learning:driving,DRL:dirivng:17}. The trend is especially fueled by the release of excellent driving simulators   
\cite{AirSim:17,CARLA:simulator}.   

\subsection{Assistive Features of Vehicle with Surround View Cameras}
Over the last decades, more and more assistive technologies have been deployed to vehicles, that increase driving safety. Technologies such as lane keeping, blind spot checking, forward collision avoidance, adaptive cruise control, driver behavior prediction etc., alert drivers about potential dangers~\cite{forecast:control,control:driver:modeling,kasper2012object,car:knows:iccv15}. Research in this vein recently has shifted focus to surround-view cameras, as a panoramic view around the vehicle is  needed for many such applications. Notable examples include object detection, object tracking, lane detection, maneuver estimation, and parking guidance. For instance, a bird's eye view has been used to monitor the surrounding of the vehicle in~\cite{bird:view:surrounding:monitoring:08}. Trajectories and maneuvers of surrounding vehicles are estimated with surround view camera arrays~\cite{trajectory:surrounding:vehicles:16,SV:trajectory:analysis:16}. Datasets, methods and evaluation metrics of object detection and tracking  with multiple overlapping cameras are studied in ~\cite{detection:360:IV:15,mpers:detection:trackingITSC:16}. Lane detection with surround-view cameras is investigated in~\cite{AV:nomitoring:13} and the parking problem in~\cite{SV:parking:iv15}. Advanced driver assistance systems often use a 3-D surround view, which informs drivers about the environment and eliminates blind spots~\cite{3d:surround:adas:18}. Our work adds autonomous driving to this list. Our dataset can also be used for all aforementioned problems; and it provides a platform to study the usefulness of route planners.

\subsection{Navigation and Maps} 
In-car navigation systems have been widely used to show the vehicle’s current location on a map and to inform drivers on how to get from the current position to the destination. Increasing the accuracy and robustness of systems for positioning, navigation and digital maps has been another research focus for many years. Several methods for high-definition mapping have been proposed~\cite{HD:map:10,HD:map:12}, some specifically for autonomous driving~\cite{traffic:rules:AV:15,Lanelets:14}. Route planning has been extensively studied as well~\cite{Driving:knowledge:world:11,route:planning,scenic:driving:route:13,personalized:TripPlanner:15,towards:personalized:routing:15}, mainly to compute the fastest, most fuel-efficient, or a customized trajectory to the destination through a road network.  
Yet, thus far their usage is mostly restricted to help human drivers. Their accessibility as an aid to learn autonomous driving models has been limited. This work reports on two ways of using two kinds of maps: a s-o-t-a commercial map TomTom Maps~\footnote{\url{https://www.tomtom.com/en_us/drive/maps-services/maps/}} and the excellent collaborative project OpenStreetMaps~\cite{openstreetmap:08}.   

While considerable progress has been made both in computer vision and in route planning, their integration for learning driving models has not received due attention in the academic community. A trending topic is to combine digital maps and street-view images for accurate vehicle localization~\cite{vehicle:localization:map:10,mapping:gps:lane:marking:13,local:align:image:map:13,map:net:17}. 

\subsection{Driving Scene Understanding}
Road scene understanding is a crucial enabler for assisted or autonomous driving. Typical examples include the detection of roads~\cite{recent:progress:lane}, traffic lights~\cite{traffic:light:survey:16}, cars and pedestrians~\cite{renNIPS15fasterrcnn,DomainAdaptiveFasterRCNN,Cityscapes,semantic:foggy:scene}, and the tracking of such objects~\cite{vehicles:road:survey:13,fastflow16,pathtrack}. We refer the reader to these comprehensive surveys~\cite{cv4av:17,looking:at:human}. Integrating recognition results like these of the aforementioned algorithms may well be necessary but is beyond the scope of this paper.  

\section{The Driving Dataset }
\label{sec:dataset}

We first present our sensor setup, then describe our data collection, and finally compare our dataset to other driving datasets.

\subsection{Sensors}
Three kinds of sensors are used for data collection in this work: cameras, a route planner (with a map), and a USB reader for data from the vehicle's CAN bus.   

\begin{figure*}[tb]
\centering
 $ \begin{array}{ccccc}  
    \includegraphics[width=0.6\linewidth]{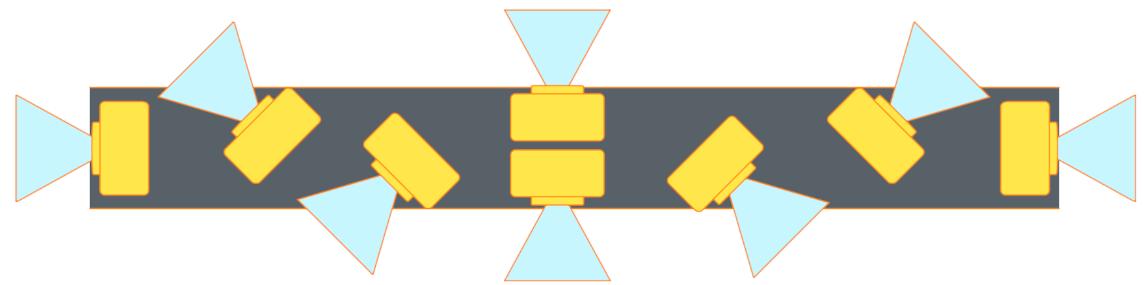} &
    \includegraphics[width=0.35\linewidth]{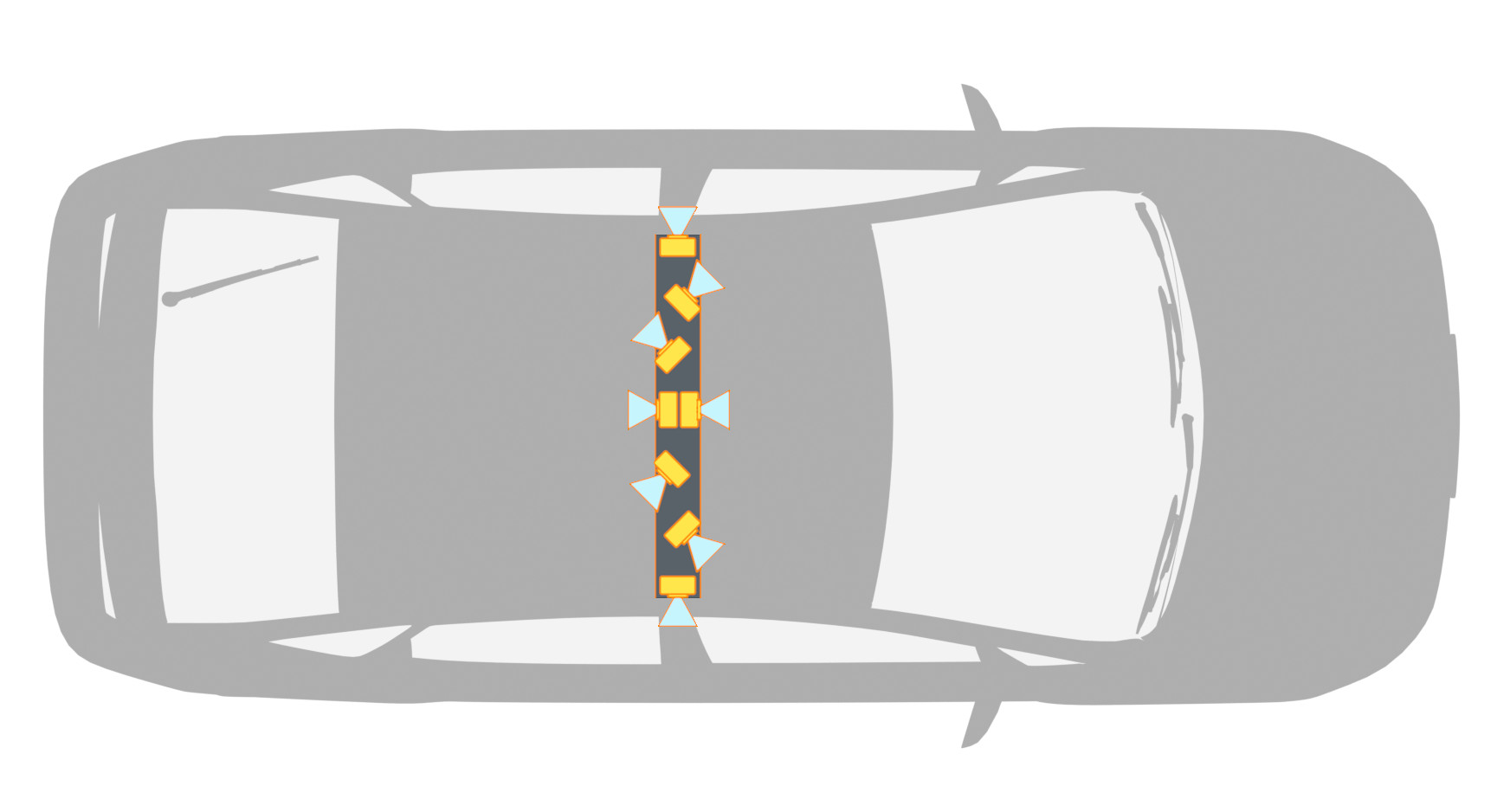} \\
   \text{(a) our camera rig} & \text{(b) rig on the  vehicle}\\
\end{array} $
   \caption{The configuration of our cameras. The rig is $1.6$ meters wide so that the side-view cameras can have a good view of road surface without the obstruction by the roof of the vehicle. The cameras are evenly distributed laterally and angularly.}
   \label{fig:rig:car}
   \end{figure*}

\noindent
\textbf{Cameras}. 
We use eight cameras and mount them on the roof of the car using a specially designed rig with 3D printed camera mounts. The cameras are mounted under the following angles: $0^o$, $45^o$, $90^o$,$135^o$, $180^o$, $225^o$, $270^o$ and $315^o$ relative to the vehicle's heading direction. We installed GoPro Hero 5 Black cameras, due to their ease of use, their good image quality when moving, and their weather-resistance. All videos are recorded at 60 frames per second (fps) in 1080p. As a matter of fact, a full 360-degree view can be covered by four cameras already. Please see Figure~\ref{fig:rig:car} for our camera configuration. 

\noindent
\textbf{Route Planners}. 
Route planners have been a research focus over many years~\cite{route:planning,scenic:driving:route:13}. While considerable progress has been made both in computer vision and in route planning, their integration for learning to drive has not received due attention in the academic community. Routing has become ubiquitous with commercial maps such as Google Maps, HERE Maps, and TomTom Maps, and on-board navigation devices are virtually in every new car. Albeit available in a technical sense, their routing algorithms and the underlying road networks are not yet accessible to the public. In this work, we exploited two route planners: one based on TomTom Map and the other on OpenStreetMap. 

TomTom Map represents one of the s-o-t-a commercial maps for driving applications. Similar to all other commercial counterparts, it does not provide open APIs to access their `raw' data. We thus exploit the visual information provided by their TomTom GO Mobile App~\cite{TomTom:mobile:go}, and recorded their rendered map views using the native screen recording software supplied by the smart phone, an iPhone 7. Since map rendering comes with rather slow updates, we capture the screen at 30 fps. The video resolution was set to $1280 \times 720$ pixels. 

Apart from the commercial maps, OpenStreetMaps (OSM)~\cite{openstreetmap:08} has gained a great attention for supporting routing services.
The OSM geodata includes detailed spacial and semantic information about roads, such as name of roads, type of roads (e.g. highway or footpath), speed limits, addresses of buildings, etc. The effectiveness of OSM for Robot Navigation has been demonstrated by Hentschel and Wagner~\cite{navigation:openstreetmap:10}. We thus, in this work, use the real-time routing method developed by Luxen and Vetter for OSM data~\cite{routing:openstreetmap:11} as our second route planner. The past driving trajectories (a stack of GPS coordinates) are provided to the routing algorithm to localize the vehicle to the road network, and the GPS tags of the planned road for the next $300$ meters ahead are taken as the representation of the planned route for the `current' position. Because the GPS tags of the road networks of OSM are not distributed evenly according to distance, we fitted a cubic smoothing spline to the obtained GPS tags and then sampled $300$ data points from the fitted spline with a stride of $1$ meter. Thus, for the OSM route planner, we have a $300 \times 2$ matrix ($300$ GPS coordinates) as the representation of the planned route for every `current' position.  

\noindent
\textbf{Human Driving Maneuvers}. 
We record low level driving maneuvers, i.c. the steering wheel angle and vehicle speed, registered on the CAN bus of the car at 50Hz. The CAN protocol is a simple ID and data payload broadcasting protocol that is used for low level information broadcasting in a vehicle. As such, we read out the specific CAN IDs and their corresponding payload for steering wheel angle and vehicle speed via a CAN-to-USB device and record them on a computer connected to the bus.

\noindent
\textbf{Vehicle's Odometry}. 
We use the GoPro cameras' built-in GPS and IMU module to record GPS data at 18Hz and IMU measurements at 200Hz while driving. This data is then extracted and parsed from the meta-track of the GoPro created video.

\subsection{Data Collection}
\noindent
\textbf{Synchronization}. 
The correct synchronization amongst all data streams is of utmost importance. For this we devised an automatic procedure that allows for synchronization to GPS for fast dataset generation. During all recording, the internal clocks of all sensors are synchronized to the GPS clock. The resulting synchronization error for the video frames is up to $8.3$ milliseconds (ms), i.e. half the frame rate. If the vehicle is at a speed of $100$ km/h, the error due to the synchronization for vehicle's longitudinal position is about 23 cm. We acknowledge that a camera which can be triggered by accurate trigger signals are preferable with respect to synchronization error. Our cameras, however, provide good photometric image quality and high frame rates, at the price of moderate synchronization error. The synchronization error of the maps to our video frame is up to $0.5$ s. This is acceptable, as the planned route (regardless of its representation) is only needed to provide a global view for navigation. The synchronization error of the CAN bus signal to our video frames is up to $10$ ms. This is also tolerable as human drivers issue driving actions at a relative low rate. For instance, the mean reaction times for unexpected and expected human drivers are
$1.3$ and $0.7$ s~\cite{driver:reaction:time:06}. 

\noindent
\textbf{\emph{Drive360} dataset}. 
With the sensors described, we collect a new dataset \emph{Drive360}. \emph{Drive360} is recorded by driving in (around) multiple cities in Switzerland. We focus on delivering realistic dataset for training driving models. Inspired by how a driving instructor teaches a human apprentice to drive, we chose the routes and the driving time with the aim to maximize the opportunity of exposing to all typical driving scenarios. This reduces the chance of generating a biased dataset with many `repetitive' scenarios, and thus allowing for an accurate judgment of the performance of the driving models. \emph{Drive360} contains $60$ hours of driving data. 

The drivers always obeyed Swiss driving rules, such as respecting the driving speed carefully, driving on the right lane when not overtaking a vehicle, leaving the required amount of distance to the vehicle in front etc. We have a second person accompanying the drivers to help (remind) the driver to always follow the route planned by our route planner. We have used a manual setup procedure to make sure that the two route planners generate the `same' planned route, up to the difference between their own representations of the road networks. After choosing the starting point and the destination, we first generate a driving route  with the OSM route planner. For TomTom route planner, we obtain the same driving route by using the same starting point and destination, and by adding a consecutive sequence of waypoints (intermediate places) on the route. We manually verified every part of the route before each driving to make sure that the two planned routes are in deed the same. After this synchronization, TomTom Go Mobile is used to guide our human drivers due to its high-quality visual information. The data for our OSM route planner is obtained by using the routing algorithm proposed in~\cite{routing:openstreetmap:11}. In particular, for each `current' location, the `past' driving trajectory is provided to localize the vehicle on the originally planned route in OSM. Then the GPS tags of the route for the next $300$ meters ahead are retrieved.  

\subsection{Comparison to other datasets}
In comparison to other datasets, see Table \ref{tab:dataset_compare}, ours has some unique characteristics. 

\noindent
\textbf{Planned routes}. Since our dataset is aimed at understanding and improving the fallacies of current end-to-end driving models, we supply map data for navigation and offer the only real-world dataset to do so. It is noteworthy that planned routes cannot be obtained by post-processing the GPS coordinates recorded by the vehicle, because planned routes and actual driving trajectories intrinsically differ. The differences between the two are resulted by the actual driving (e.g. changing lanes in road construction zones and overtaking a stopped bus), and are indeed the objectives meant to be learned by the driving models. 

\noindent
\textbf{Surround views and low-level driving maneuvers}.
Equally important, our dataset  is the only dataset working with real data and offering surround-view videos with low-level driving maneuvers (e.g. steering angle and speed control). This is particularly valuable for end-to-end driving as it allows the model to learn correct steering for lane changes, requiring `mirrors' when carried out by human drivers, or correct driving actions for making turns at intersections. Compared with BDDV~\cite{end:driving:16} and Oxford dataset~\cite{oxford:driving}, we offer low level driving maneuvers of the vehicle via the CAN bus, whereas they only supply the cars ego motion via GPS devices. This allows us to predict input control of the vehicle which is one step closer to a fully autonomous end-to-end trained driving model. Udacity~\cite{udacity} also offers low-level driving maneuvers via the CAN bus. It, however, lacks of route planners and contains only a few hours of driving data. 

\noindent
\textbf{Dataset focus}. 
As shown in Table~\ref{tab:dataset_compare}, there are multiple datasets compiled for tasks relevant to autonomous driving. These datasets, however, all have their own focuses. KITTI, Cityscapes and GTA focus more on semantic and geometric understanding of the driving scenes. Oxford dataset focus on capturing the temporal (seasonal) changes of driving scenes, and thus limited the driving to a `single' driving route. BDDV~\cite{end:driving:16} is a very large dataset, collected from many cities in a crowd-sourced manner. It, however, only features a front-facing dashboard camera. 

\setlength{\tabcolsep}{2.4pt}
\rowcolors{2}{}{gray!10}
\begin{table*}[tb]
\centering
\footnotesize
\rowcolors{2}{gray!50}{white}
  \begin{tabular}{ccccccccccc}
   \toprule
  \rotatebox[origin=c]{90}{Datasets}  &  \rotatebox[origin=c]{90}{\thead{driving \\ time (h)}} & \rotatebox[origin=c]{90}{\# cams} & \rotatebox[origin=c]{90}{fps} & \rotatebox[origin=c]{90}{\thead{maneuvers, \\ e.g. steering}} & \rotatebox[origin=c]{90}{\thead{route \\ planner}} & \rotatebox[origin=c]{90}{\thead{GPS\\IMU}} & \rotatebox[origin=c]{90}{\thead{control of \\ cam pose}} & \rotatebox[origin=c]{90}{data type} & \rotatebox[origin=c]{90}{lidar} \\    \midrule
   \ourdata & 60  & 8, \ominicam \ominicamrotated & 60 & \YES & \YES & \YES & \YES & real  & \ding{55} \\   
     KITTI \cite{kitti:dataset} & 1  & 2,  \fcam & 10 & \ding{55} & \ding{55}& \YES & \YES & real & \YES\\ 
     Cityscapes \cite{Cityscapes} & $<100$ & 2, \fcam & 16 & \ding{55} & \ding{55}& \YES & \YES & real & \ding{55} \\ 
     Comma.ai & 7.3 & 1, \fcam & 20 & \YES & \ding{55}& \YES & N.A. & real & \ding{55} \\ 
     Oxford \cite{oxford:driving} & 214 & 4, \ominicam & 16 & \ding{55} & \ding{55}& \YES & \YES & real & \YES \\ 
     BDDV \cite{end:driving:16} & 10k & 1,  \fcam & 30 & \ding{55} & \ding{55}& \YES & \ding{55} & real & \ding{55} \\ 
     Udacity \cite{udacity} & 1.1 & 3, \udacitycam & 30 & \YES & \ding{55}& \YES & N.A. & real & \ding{55}\\ 
     GTA & N.A. & 1  \fcam & \YES & \YES & \ding{55} & N.A. & rendered & synthetic & \ding{55} \\ 
     \bottomrule
  \end{tabular}
  \caption{Comparison of our dataset to others compiled for driving tasks (cam=camera).}
  \label{tab:dataset_compare}
 \end{table*}


\section{Approach}
\label{sec:approach} 
The goal of our driving model is to map directly from the planned route, the historical vehicle states and the current road situations, to the desired driving actions. 
\subsection{Our Driving Model} 
\label{sec:problem}

Let us denote by $\mathrm{I}$ the surround-view video, $\mathrm{P}$ the planned route, $\mathrm{L}$ the vehicle's location, and $\mathrm{S}$ and $\mathrm{V}$ the vehicle's steering angle and speed. We assume that the driving model works with discrete time and makes driving decisions every $1/f$ seconds. The inputs are all synchronized and sampled at sampling rate $f$.  Unless stated otherwise, our inputs and outputs all are represented in this discretized form. 

We use subscript $t$ to indicate the time stamp. For instance, the current video frame is $I_t$, the current vehicle's speed is $V_t$, the $k^{th}$ previous video frame is $I_{t-k}$, and the $k^{th}$ previous steering angle is $S_{t-k}$, etc. Then, the $k$ recent samples can be denoted by $\mathbf{V}_{[t-k+1,t]} \equiv \langle V_{t-k+1}, ..., V_{t}\rangle$, $\mathbf{S}_{[t-k+1,t]} \equiv \langle S_{t-k+1}, ..., S_{t}\rangle$ and $\mathbf{V}_{[t-k+1,t]} \equiv \langle V_{t-k+1}, ..., V_{t}\rangle$, respectively. 
Our goal is to train a deep network that predicts desired driving actions from the  vehicle's historical states, historical and current visual observations, and the planned route. The learning task can be defined as: 
\begin{equation}
F: (\mathcal{S}_{[t-k+1,t]}, \mathcal{V}_{[t-k+1,t]}, \mathcal{L}_{[t-k+1,t]}, \mathcal{I}_{[t-k+1,t]}, P_t) \rightarrow \mathcal{S}_{t+1} \times \mathcal{V}_{t+1}
\label{eq:learning:fun1}
\end{equation}
where $\mathcal{S}_{t+1}$ represents the steering angle space and $\mathcal{V}_{t+1}$ the speed space for future time $t+1$.  
$\mathcal{S}$ and $\mathcal{V}$ can be defined at several levels of granularity. We consider the continuous values directly recorded from the car's CAN bus, where $\mathcal{V}=\{V | 0  \leq V \leq 180$ for speed and $\mathcal{S}=\{S | -720  \leq S \leq 720\}$ for steering angle. Here, kilometer per hour (km/h) is the unit of $V$, and degree ($^\circ$) the unit of $S$. Since there is not much to learn from the historical values of $\mathrm{P}$, only $P_t$ is used for the learning. $P_t$ is either a video frame from our TomTom route planner or a $300 \times 2$ matrix from our OSM route planner. 


Given $N$ training samples collected during real drives, learning to predict the driving actions for the future time $t+1$ is based on minimizing the following cost: 
\begin{equation}
\begin{split}
L(\theta) = \sum_{n=1}^N \Big( l({S}^{n}_{t+1}, F_{\text{s}}(\mathbf{S}^{n}_{[t-k+1, t]}, \mathbf{V}^{n}_{[t-k+1, t]}, \mathbf{L}^{n}_{[t-k+1, t]}, \mathbf{I}^{n}_{[t-k+1, t]}, P_t)) \\ + \lambda  l({V}^{n}_{t+1}, F_{\text{v}}(\mathbf{S}^{n}_{[t-k+1, t]}, \mathbf{V}^{n}_{[t-k+1, t]}, \mathbf{L}^{n}_{[t-k+1, t]}, \mathbf{I}^{n}_{[t-k+1, t]}, P_t))\Big), 
\end{split}
\end{equation}
where $\lambda$ is a parameter balancing the two losses, one for steering angle and the other for speed. We use $\lambda=1$ in this work. $F$ is the learned function for the driving model. For the continuous regression task, $l(.)$ is the $L2$ loss function. Finding a better way to balance the two loss functions constitutes our future work. Our model learns from multiple previous frames in order to better understand traffic dynamics. 

\subsection{Implementation} 
\label{sec:networkmodel}

Our driving system is trained with four cameras (front, left, right, and rear view), which provide a full panoramic view already. We recorded the data with all eight cameras in order to keep future flexibility. 

This work develops a customized network architecture for our learning problem defined in Section~\ref{sec:problem}, which consists of deep hierarchical sub-networks. It comes with multiple CNNs as feature encoders, four LSTMs as temporal encoders for information from the four surround-view cameras, a fully-connected network (FN) to fuse information from all cameras and the map, and finally two FNs to output future speed and steering angle of the car. The illustrative architecture is show in Figure~\ref{fig:pipeline}.       

During training, videos are all resized to $256 \times 256$ and we augment our data by using $227 \times 227 $ crops, without mirroring. For the CNN feature encoder, we take ResNet34~\cite{resnet} model pre-trained on the ImageNet~\cite{imagenet} dataset. Our network architecture is inspired by the Long-term Recurrent Convolutional Network developed in \cite{lrcn2014}. A more detailed description about the network architecture is provided in the supplementary material. 

\section{Experiments}
\label{sec:experiment} 
\noindent
We train our models on 80\% of our dataset, corresponding to 48 hours of driving time and around 1.7 million unique synchronized sequence samples. Our driving routes are normally $2$ hours long. We have selected $24$ out of the $30$ driving routes for training, and the other $6$ for testing. This way, the network would not overfit to any type of specific road or weather. 
Synchronized video frames are extracted at a rate of $10$ fps, as $60$ fps will generate a very large dataset. A synchronized sample contains four frames at a resolution of $256\times 256$ for the corresponding front, left, right and rear facing cameras, a rendered image at $256 \times 256$ pixels for TomTom route planner or a $300 \times 2$ matrix for OSM route planner, CAN bus data and the GPS data of the the `past'. 

We train our models using the Adam Optimizer with an initial learning rate of $10^{-4}$ and a batch size of 16 for 5 epochs, resulting in a training time of around 3 days. For the four surround-view cameras, we have used four frames to train the network: 0.9s in the past, 0.6s in the past, 0.3s in the past, and the current frame. This leads to a sampling rate of $f=3.33$. A higher value can be used at the price of computational cost. This leads to $4 \times 4 = 16$ CNNs for capturing street-view visual scene. 
  
We structure our evaluation into two parts: evaluating our method against existing methods, and evaluating the benefits of using a route planner and/or a surround-view camera system.

\subsection{Comparison to other single-camera methods}
\label{lab:arch_eval}
\noindent
We compare our method to the method of \cite{end:driving:16} and \cite{nvidia:driving:16}. Since BDDV dataset does not provide data for driving actions (e.g. steering angle)~\cite{end:driving:16}, we train their networks on our dataset and compare with our method directly. For a fair comparison, we follow their settings, by only using a single front-facing camera and predicting the driving actions for the future time at $0.3s$.

\setlength{\tabcolsep}{3pt}
\begin{SCtable}[][tb]
\centering
\begin{tabular}[scale=0.5]{ccccccccc}
\toprule
 &  CAN-only & \cite{nvidia:driving:16} &\cite{end:driving:16} & Ours \\
\hline
Steering & 0.869 & 1.312 & 0.161 & \textbf{0.134} \\
\hline
Speed & 0.0147 & 0.6533 & 0.0066 & \textbf{0.0030} \\
\bottomrule
\end{tabular}
\caption{MSE of speed prediction and steering angle prediction when a single front-facing camera is used (previous driving states are given).}
\label{tab:eval1}
\end{SCtable}

We use the mean squared error (MSE) for evaluation. The results for speed prediction and steering angle prediction are shown in Table~\ref{tab:eval1}. We include a baseline reference of only training on CAN bus information (no image information given).
The table shows that our method outperforms \cite{nvidia:driving:16} significantly and is slightly better than \cite{end:driving:16}. \cite{nvidia:driving:16} does not use a pre-trained CNN; this probably explains why their performance is a lot worse. The comparison to these two methods is to verify that our frontal-view driving model represents the state of the art so that the extension is made to a sensible basis to include multiple-view cameras and to include route planners.

We note that the baseline reference performs quite well, suggesting that due to the inertia of driving maneuvers, the network can already predict speed and steering angle of $0.3s$ further into the future quite well, solely based on the supplied ground truth maneuver of the past. For instance, if one steers the wheels to the right at time $t$, then at $t+0.3s$ the wheels are very likely to be at a similar angle to the right. In a true autonomous vehicle the past driving states might not be always correct. Therefore, we argue that the policy employed by some existing methods by relying on the past `ground-truth' states of the vehicle should be used with caution. For the real autonomous cars, the errors will be exaggerated via a feedback loop. Based on this finding, we remove $\mathcal{S}_{[t-k+1,t]}$ and $\mathcal{V}_{[t-k+1,t]}$, i.e. without using the previous human driving maneuvers, and learn the desired speed and steering angle only based on the planned route, and the visual observations of the local road situation.  This new setting `forces' the network to learn knowledge from route planners and road situations. 


\subsection{Benefits of Route Planners}
\label{sec:routeplanner}
\noindent
We evaluate the benefit of a route planner by designing two networks using either our visual TomTom, or our numerical OSM guidance systems, and  compare these against our network that does not incorporate a route planner. 
The results of each networks speed and steering angle prediction are summarized in Table~\ref{tab:eval2_angle}. The evaluation shows that our visual TomTom route planner significantly improves prediction performance, while the OSM approach does not yield a clear improvement. Since, the prediction of speed is easier than the prediction of steering angle, using a route planner will have a more noticeable benefit on the prediction of steering angles.

\setlength{\tabcolsep}{8pt}
\begin{table} [tb]
\centering
\begin{tabular}{ccccccc} \hline
\multirow{2}{*}{Cameras} & \multirow{2}{*}{Route planner} & \multicolumn{2}{c}{Full dataset} & \multicolumn{2}{c}{Subset: GT $\leq 30$ km/h} \\ \cline{3-6}
 \multicolumn{2}{l}{} & Steering & Speed & Steering &  Speed \\  \hline
\multirow{3}{*}{Front-view} & None & 0.967 & 0.197 & 4.053 & 0.167  \\
&  TomTom & 0.808 & \textbf{0.176} & 3.357 & 0.268  \\
&  OSM & 0.981 & 0.212 & 4.087 & 0.165  \\ 
\cline{1-6} 
\multirow{3}{*}{Surround-view} & None & 0.927 & 0.257 & 3.870 & \textbf{0.114} \\
&  TomTom & \textbf{0.799} & 0.200 & \textbf{3.214} & 0.142 \\
&  OSM & 0.940 & 0.228 & 3.917 & 0.125 \\ \hline
\end{tabular}
\caption{MSE (smaller=better) of speed and steering angle prediction by our method, when different settings are used. Predictions on full evaluation set and the subset with human driving maneuver $\leq 30$ km/h. }
\label{tab:eval2_angle}
\end{table}


\noindent
\textbf{Why the visual TomTom planner is better?}
It is easy to think that GPS coordinates contain more accurate information than the rendered videos do, and thus provide a better representation for planned routes. This is, however, not case if the GPS coordinates are used directly without further, careful, processing. The visualization of a planned route on navigation devices such as TomTom Mobile Go makes use of accurate vehicle localization based on vehicle's moving trajectories to provide accurate procedural knowledge of the routes along the driving direction. The localization based on vehicle's moving trajectories is tackled under the name \emph{map-matching}, and this, in itself, is a long-standing research problem~\cite{map:low-sampling-rate:gps:09,interactive:map:matching:10,map:matching:14}. For our TomTom route planner, this is done with TomTom's excellent underlying \emph{map-matching} method, which is unknown to the public though. This rendering process converts the `raw' GPS coordinates into a more structural representation. Our implemented OSM route planner, however, encodes more of a global spatial information at a map level, making the integration of navigation information and street-view videos more challenging. Readers are referred to Figure~\ref{fig:examples:pics} for exemplar representations of the two route planners.

In addition to \emph{map-matching}, we provide further possible explanations: \textbf{1)} raw GPS coordinates are accurate for locations, but fall short of other high-level and contextual information (road layouts, road attributes, etc.) which is `visible' in the visual route planner. For example, raw GPS coordinates do not distinguish `highway exit' from `slight right bend' and do not reveal other alternative roads in an intersection, while the visual route planner does. 
It seems that those semantic features optimized in navigation devices to assist human driving are useful for machine driving as well. Feature designing/extraction for the navigation task of autonomous driving is an interesting future topic.  \textbf{2)} The quality of underlying road networks are different from TomTom to OSM. OSM is crowdsourced, so the quality/accuracy of its road networks is not always guaranteed. It is hard to make a direct comparison though, as TomTom's road networks are inaccessible to the public.

\subsection{Benefits of  Surround-View Cameras}
\label{sec:mcam}
\noindent
Surround-view cameras offer a modest improvement for predicting steering angle on the full evaluation set. They, however, appear to reduce the overall performance for speed prediction. Further investigation has shown that surround-view cameras are especially useful for situations where the ego-car is required to give the right of way to other (potential) road users by controlling driving speed. Notable examples include 1) busy city streets and residential areas where the human drives at low velocity; and 2) intersections, especially those without traffic lights and stop signs. For instance, the speed at an intersection is determined by whether the ego-car has a clear path for the planned route. Surround-view cameras can see if other cars are coming from any side, whereas a front camera only is blind to many directions. In order to examine this, we have explicitly selected two specific types of scenes across our evaluation dataset for a more fine-grained evaluation of front-view vs. surround-view: 1) low-speed (city) driving according to the speed of human driving; and 2)  intersection scenarios by human annotation. The evaluation results are shown in Table~\ref{tab:eval2_angle} and Table~\ref{tab:low_speed_eval}, respectively. The better-performing TomTom route planner models are used for the experiments in Table~\ref{tab:low_speed_eval}. Surround-view cameras significantly improve the performance of speed control in these two very important driving situations. For `high-speed' driving on highway or countryside road, surround-view cameras do not show clear advantages, in line with human driving -- human drivers also consult non-frontal views less frequently for high-speed driving. 

\setlength{\tabcolsep}{8pt}
\begin{table} [tb]
\centering
\begin{tabular}{ccccccc} \hline
Cameras & $\leq 10$ km/h & $\leq 20$ km/h & $\leq 30$ km/h & $\leq 40$ km/h & $\leq 50$ km/h \\ \hline
Front-view & 0.118 & 0.150 & 0.158 & 0.157 & 0.148 \\
Surround-view & \textbf{0.080} & \textbf{0.127} & \textbf{0.145} & \textbf{0.146} & \textbf{0.143} \\
\hline
\end{tabular}
\caption{MSE (smaller=better) of speed prediction by our Front-view+TomTom and Surround-view+TomTom driving models. Evaluated on manually annotated intersection scenarios over a 2-hour subset of our evaluation dataset. Surround-view significantly outperforms front-view in intersection situations.}
\label{tab:low_speed_eval}
\end{table}

As a human driver, we consult our navigation system mostly when it comes to multiple choices of road, namely at road intersections. To evaluate whether route planning improves performance specifically in these scenarios, we select a subset of our test set for examples with a low speed  by human, and report the results in this subset also in Table \ref{tab:eval2_angle}. Results in Table \ref{tab:eval2_angle} supports our claim that route planning is beneficial to a driving model, and improves the driving performance especially for situations where a turning maneuver is performed. In future work, we plan to select other interesting situations for more detailed evaluation.

\subsubsection{Qualitative Evaluation}
\label{sec:extreme_case_eval}
While standard evaluation techniques for neural networks such as mean squared error, do offer global insight into the performance of models, they are less intuitive in evaluating where, at a local scale, using surround view cameras or route planning improves prediction accuracy. 
To this end, we use our visualization tool to inspect and evaluate the model performances for different `situations'.

Figure \ref{fig:examples:pics} shows examples of three model comparisons (TomTom, Surround, Surround+TomTom) row-wise, wherein the model with additional information is directly compared to our front-camera-only model, shown by the speed and steering wheel angle gauges. The steering wheel angle gauge is a direct map of the steering wheel angle to degrees, whereas the speed gauge is from 0km/h to 130km/h. Additional information a model might receive is `image framed' by the respective color. Gauges should be used for relative model comparison, with the front-camera-only model prediction in orange, model with additional information in red and human maneuver in blue. Thus, for our purposes, we define a well performing model when the magnitude of a model gauge is identical (or similar) to the human gauge. Column-wise we show examples where: (a) both models perform well, (b) model with additional information outperforms , (c) both models fail.

\setlength{\tabcolsep}{1pt}
\begin{figure*}[!tb]
$\begin{tabular}{cccc}
\text{(1)} &
\includegraphics[width=0.30\linewidth]{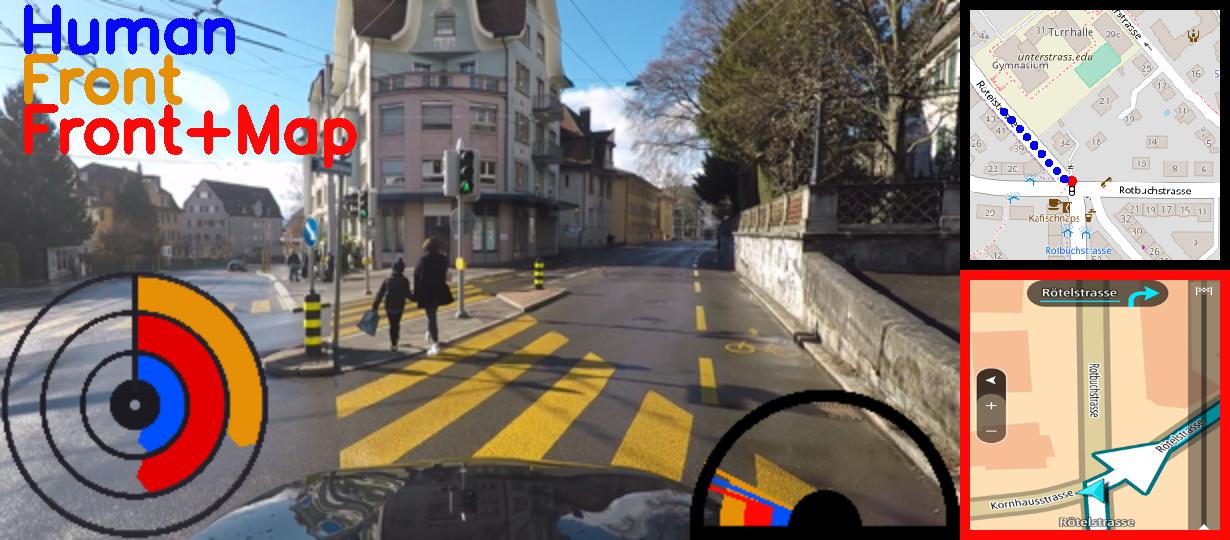}
&
\includegraphics[width=0.30\linewidth]{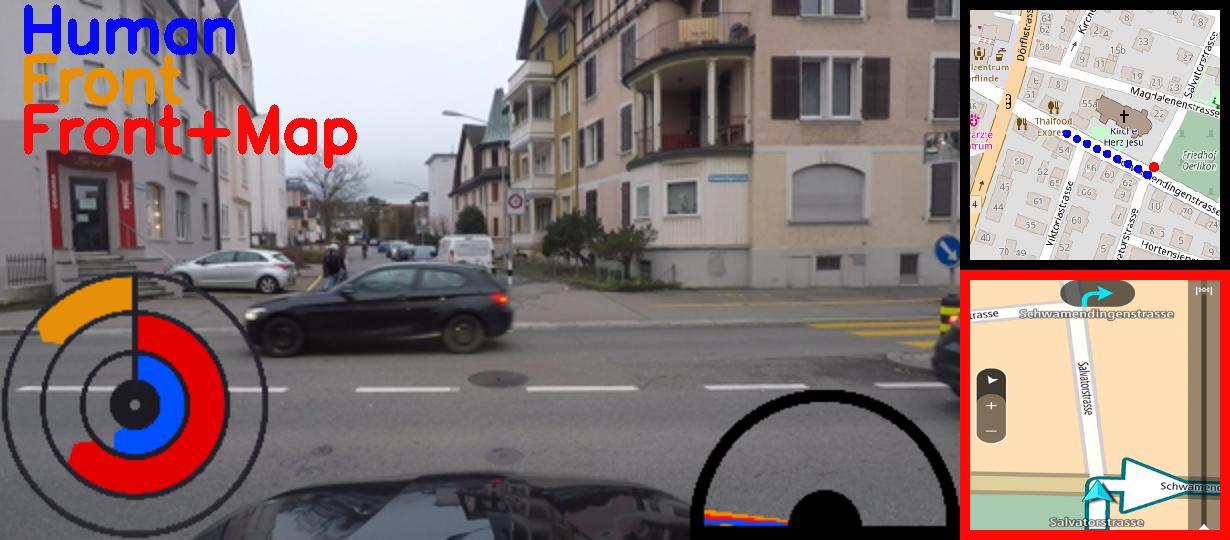}
&
\includegraphics[width=0.30\linewidth]{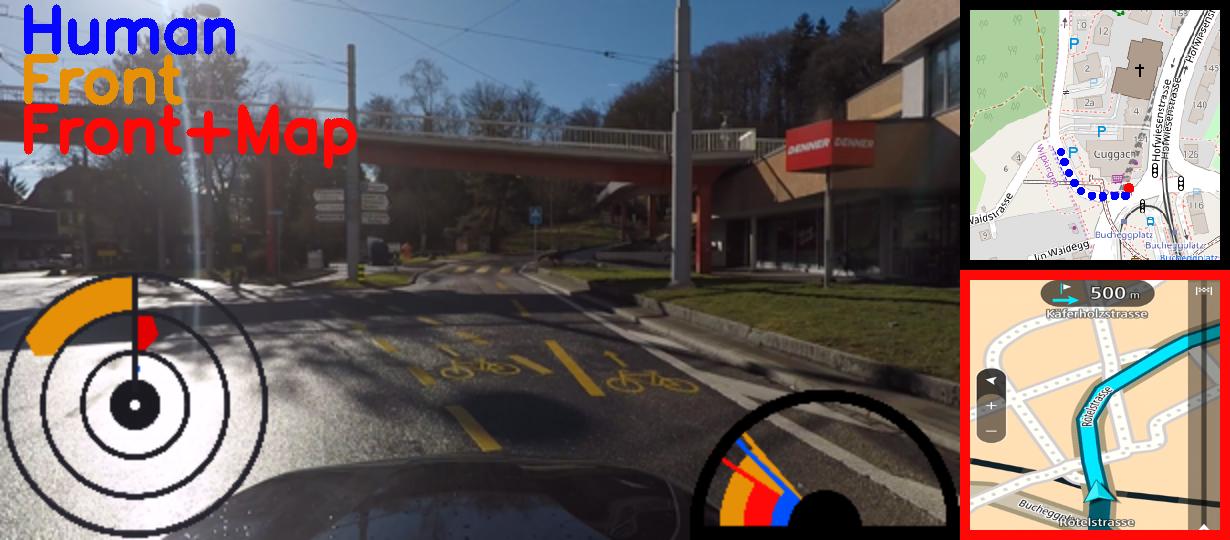} \\ 
\text{ (2)} &
\includegraphics[width=0.30\linewidth]{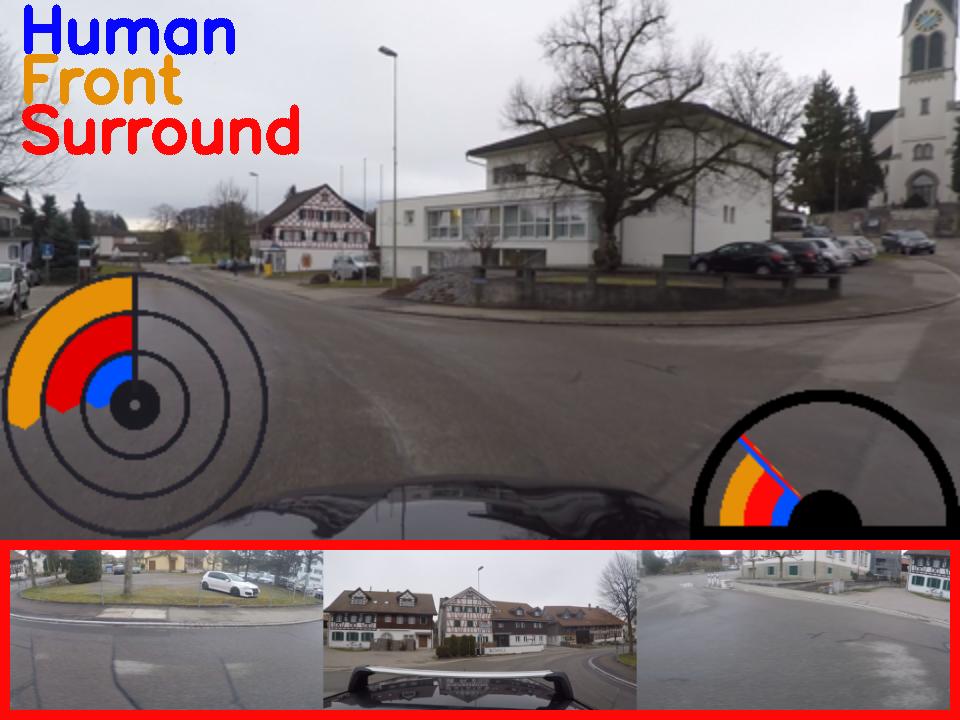}
&
\includegraphics[width=0.30\linewidth]{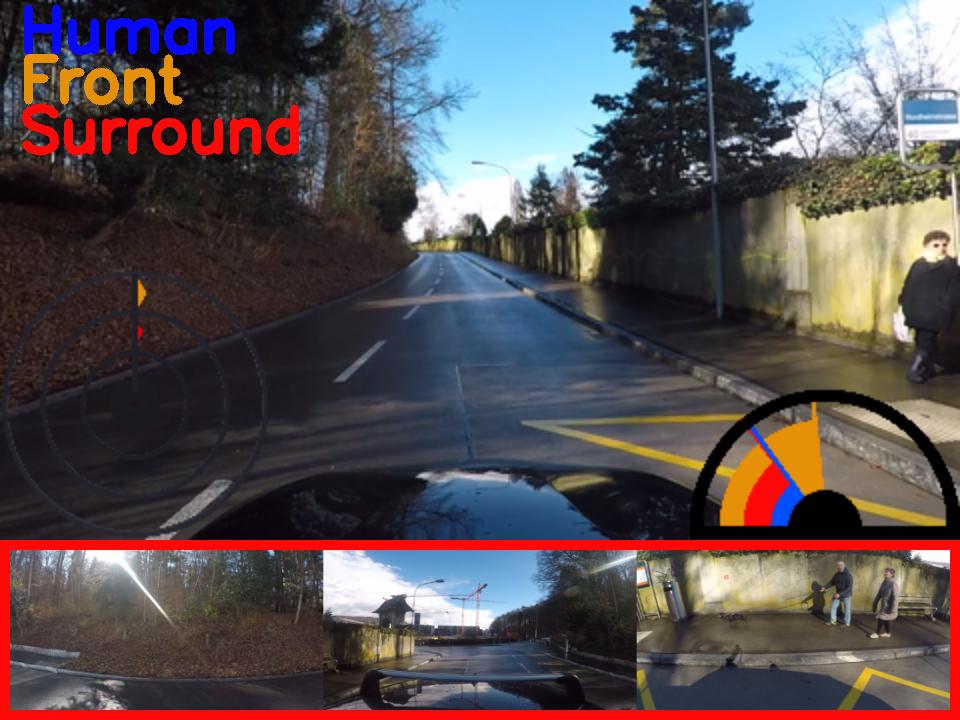}
& 
\includegraphics[width=0.30\linewidth]{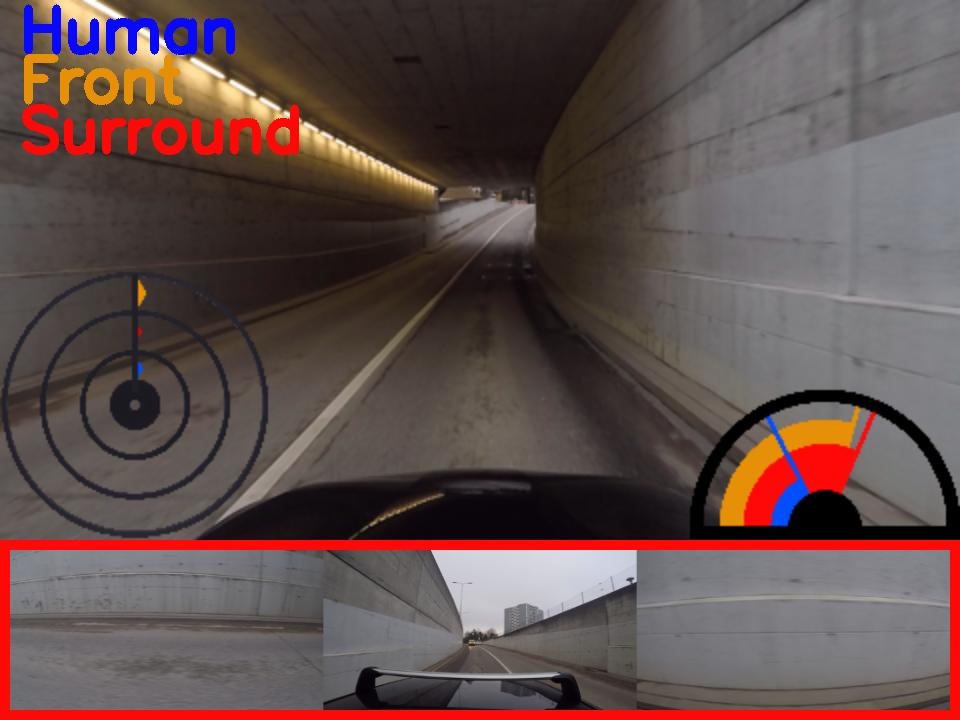} \\ 
\text{ (3)} &
\includegraphics[width=0.30\linewidth]{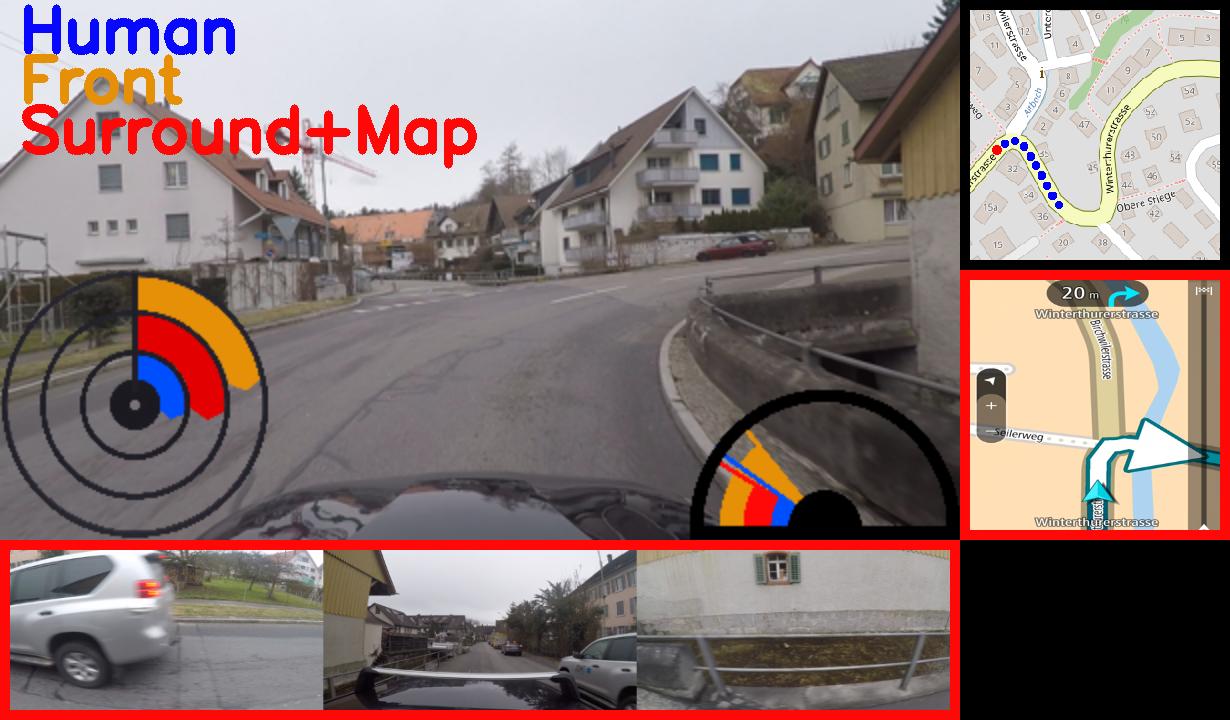}
& 
\includegraphics[width=0.30\linewidth]{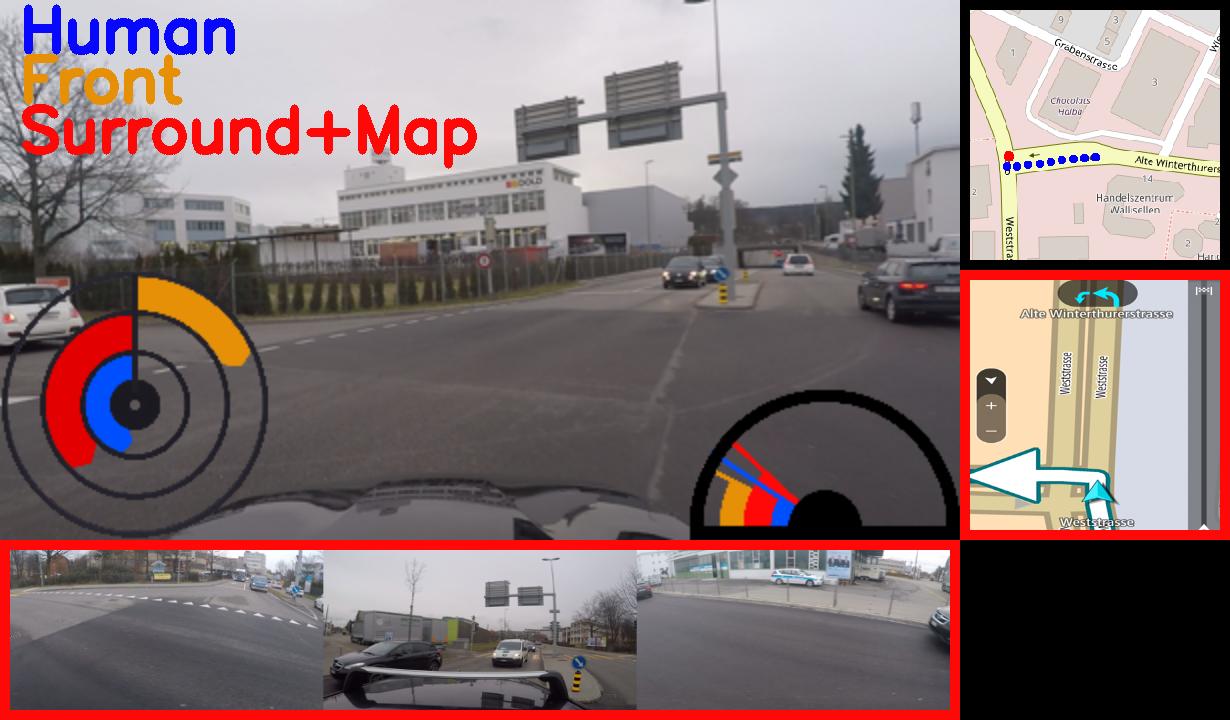}
& 
\includegraphics[width=0.30\linewidth]{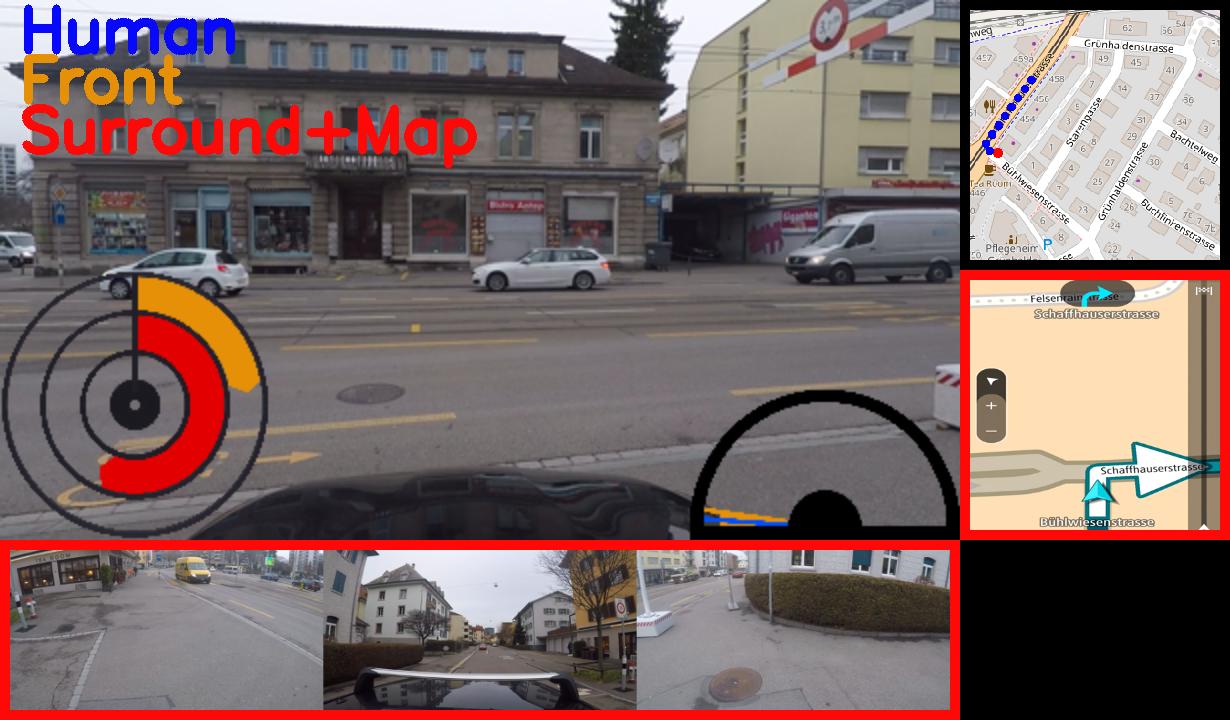} \\
& \text{ (a)} &  \text{ (b)} & \text{ (c) }
\end{tabular}$
 \caption{Qualitative results for future driving action prediction, to compare three cases to the front camera-only-model: (1) learning with TomTom route planner, (2) learning with surround-view cameras (3) learning with TomTom route planner and surround-view cameras. TomTom route planer and surround-view images shown in red box, while OSM route planner in black box.  Better seen on screen.} 
\label{fig:examples:pics} 
\end{figure*}

Our qualitative results, in Figure \ref{fig:examples:pics} (1,b) and (3,b), support our hypothesis that a route planner is indeed useful at intersections where there is an ambiguity with regards to the correct direction of travel. Both models with route planning information are able to predict the correct direction at the intersection, whereas the model without this information predicts the opposite. While this `wrong' prediction may be a valid driving maneuver in terms of safety, it nonetheless is not correct in terms of arriving at the correct destination. Our map model on the other hand is able to overcome this. 
Figure \ref{fig:examples:pics} (2,b) shows that surround-view cameras are beneficial at predicting the correct speed. The frontal view supplied could suggest that one is on a country road where the speed limit is significantly higher than in the city, as such, our front-camera-only model predicts a speed much greater than the human maneuver. However, our surround-view system can pick up on the pedestrians on the right of the car, thus adjusts the speed accordingly. The surround-view model thus has a more precise understanding of its surroundings.

\noindent
\textbf{Visualization tool}.
To obtain further insights into where current driving models perform well or fail, we have developed a visual evaluation tool that lets users select scenes in the evaluation set by clicking on a map, and then rendering the corresponding 4 camera views, the ground truth and predicted vehicle maneuver (steering angle and speed) along with the map at that point in time. These evaluation tools along with the dataset will be released to the public. In particular, visual evaluation is extremely helpful to understand where and why a driving model predicted a certain maneuver, as sometimes, while not coinciding with the human action, the network may still predict a safe driving maneuver.

\section{Conclusion} 
\label{sec:conclusion} 
In this work, we have extended learning end-to-end driving models to a more realistic setting from only using a single front-view camera. We have presented a novel task of learning end-to-end driving models with surround-view cameras and rendered maps, enabling the car to `look' to side, rearward, and to `check' the driving direction. We have presented two main contributions: 
1) a new driving dataset, featuring $60$ hours of driving videos with eight surround-view cameras, low-level driving maneuvers recorded via car's CAN bus, two representations of planned routes by two route planners, and GPS-IMU data for the vehicle's odometry; 2) a novel deep network to map directly from the sensor inputs to future driving maneuvers. Our data features high temporary resolution and $360$ degree view coverage, frame-wise synchronization, and diverse road conditions, making it ideal for learning end-to-end driving models. Our experiments have shown that an end-to-end learning method can effectively use surround-view cameras and route planners. The rendered videos outperforms a stack of raw GPS coordinates for representing planned routes. 

\noindent
\textbf{Acknowledgements.} This work is funded by Toyota Motor Europe via the research project TRACE-Z\"urich. One Titan X used for this research was donated by NVIDIA.
\clearpage
\bibliographystyle{splncs04}
\bibliography{egbib}

\clearpage

\setcounter{equation}{0}
\setcounter{figure}{0}
\setcounter{table}{0}
\setcounter{page}{1}
\setcounter{section}{0}
\renewcommand{\theequation}{S\arabic{equation}}
\renewcommand{\thefigure}{S\arabic{figure}}

{\Large Supplementary Material}
\section{Introduction}
\label{sec:introduction}
\begin{figure*}[b]
\centering
 $ \begin{array}{ccccc}  
    \includegraphics[width=0.49\linewidth]{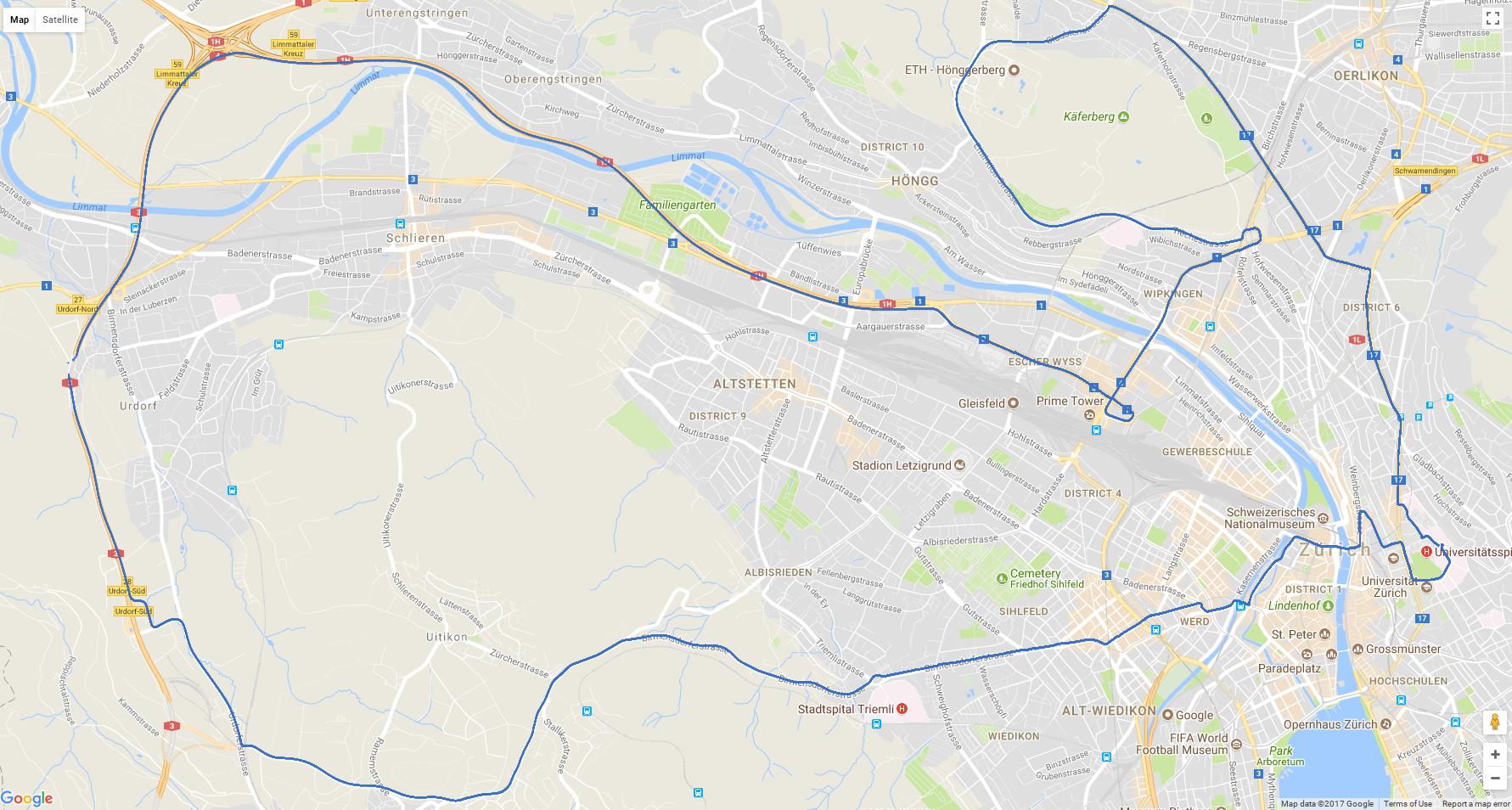} &
    \includegraphics[width=0.49\linewidth]{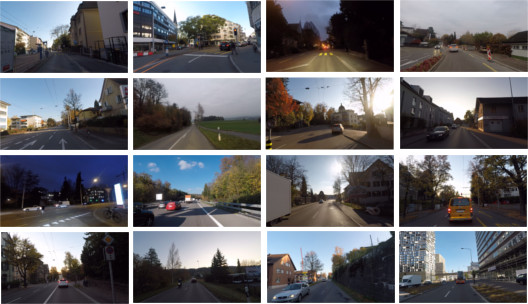} \\
     \text{ (a) a driving route} & \text{(b) image samples} \\
\end{array} $
   \caption{An example of our driving route, shown in (a), and some image examples along the driving route by our front-facing camera, shown in (b). The driving route contains varying types of roads, such as urban streets, mountainous roads, and highway.}
   \label{fig:drive_routes} 
   \end{figure*}

The supplemental material will give a detailed \textbf{architecture} description of our model (Surround-View + TomTom Route Planner), some samples from our Drive360 dataset, see Figure \ref{fig:drive_routes}, a brief qualitative evaluation study, see Figure \ref{fig:examples:pics}, and an introduction into our supplied \textbf{video}. A link to the video can be found at \url{http://people.ee.ethz.ch/~heckers/Drive360/}

\section{Architecture}
\label{sec:arch}
\begin{figure*}[t]
\centering
\includegraphics[width=0.8\linewidth]{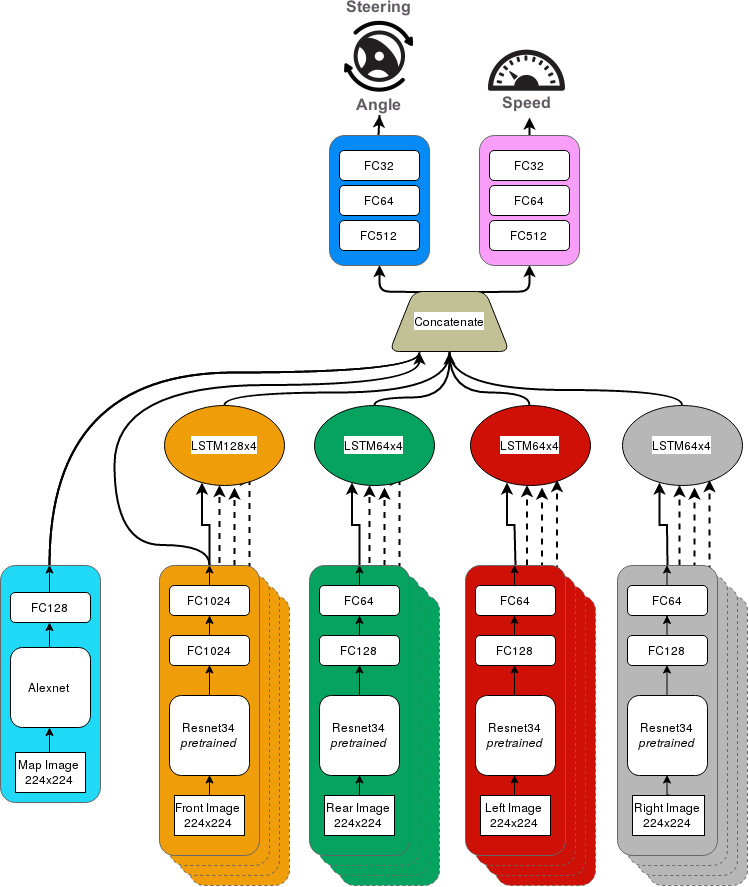}
\caption{The architecture of our Surround-View and TomTom route planner model.} 
\vspace{-6mm}
\label{fig:arch}
\end{figure*}

Figure \ref{fig:arch} illustrates our architecture for our Surround-View + TomTom Route Planner model. 

\noindent
A temporal image sequence of four sampled at 0.9s in the past, 0.6s in the past, 0.3s in the past, and the current frame is input into a pre-trained Resnet34 \cite{resnet} for each of the four surround-view cameras (front, rear, left, right). Following two fully connected layers (FC) of size 1024, the temporal feature vectors (FV) are input into a four layer LSTM with a hidden size of 128. 

\noindent
A TomTom route planner FV is extracted using Alexnet \cite{alexnet} and a single FC of size 128.

\noindent
The temporal FV, route planner FV and the front camera's current frame FV are concatenated and used as input into two FC regressor components predicting the steering wheel angle and vehicle speed at a time 0.3 seconds into the future.  

\noindent
We attribute our performance gains over \cite{end:driving:16} mainly due to the upgraded visual perception component. In particular changing the Alexnet architecture to Resnet34 to encode the surround-view camera images is an important factor.

\section{Video}
Our video has two sections. First we show our click-based visualization tool, followed by a driving model comparison between our front-camera-no-route-planner (Front) and our surround-view + TomTom route planner (Surround+TomTom) model.

\noindent
\textbf{Visualization Tool:} This  tool renders the traversed route of the car onto a map, and allows the user to select points of interest for which the local camera frames, along with the model predictions at that point, are visualized. Using this tool, we are able to quickly analyze relative model performance for 'rarer' cases such as intersections. 

\noindent
\textbf{Model comparison: } We show five driving sequences comparing the human maneuver to our Front and our Surround+TomTom model predictions. The Front model lacks multiple cameras and a route planner.

\noindent
\textit{Country Road:} both models can accurately predict the correct maneuver on a slightly right turning country road. 

\noindent
\textit{Village:} both models can navigate a more challenging sequence of following a winding road in a village setting.

\noindent
\textit{Right turn:} our Surround+TomTom model is able to anticipate the right turn maneuver, whereas the Front model only commences the turn once the vehicle has significantly turned. 

\noindent
\textit{Left turn:} our Surround+TomTom model is able to anticipate the left turn maneuver, it reduces the speed and engages in a left turn. The Front model predicts a continuing straight maneuver.

\noindent
\textit{Roundabout with Pedestrians:} both models can adjust their speed to the pedestrians crossing the road. Our Surround+TomTom model is able to take the correct, second exit out of the roundabout, whereas our Front model predicts taking the first exit, due to no route planning information. 

\setlength{\tabcolsep}{1pt}
\begin{figure*}[!tb]
$\begin{tabular}{cccc}
\includegraphics[width=0.50\linewidth]{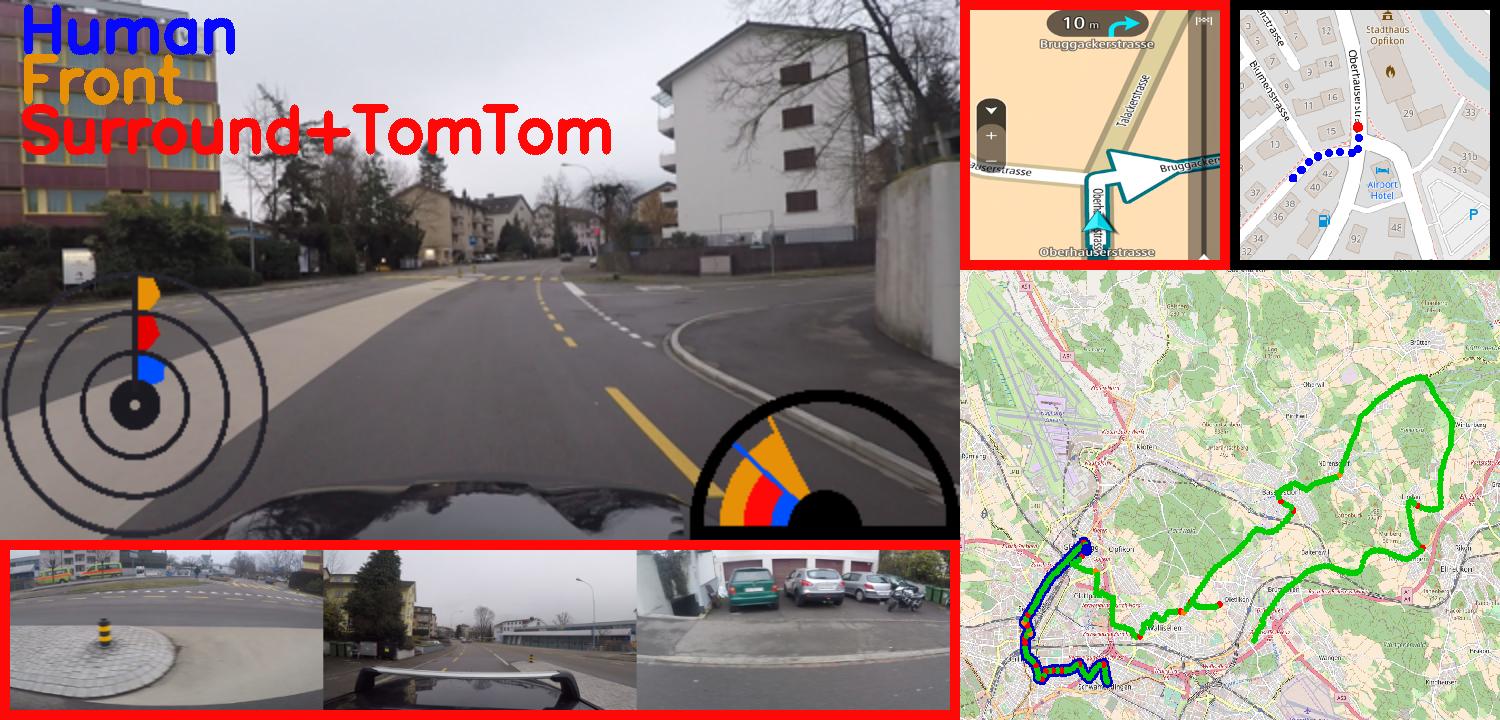}
\includegraphics[width=0.50\linewidth]{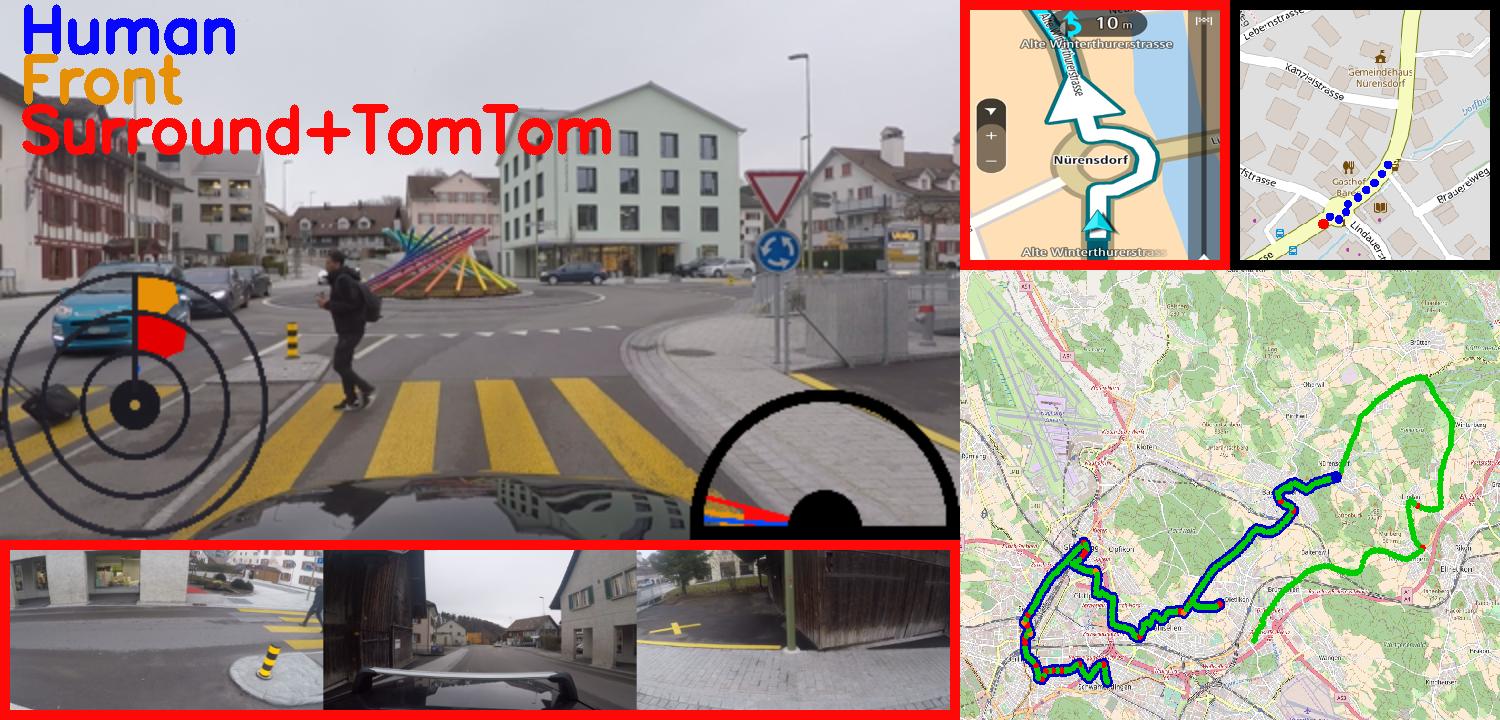} \\ 
\includegraphics[width=0.50\linewidth]{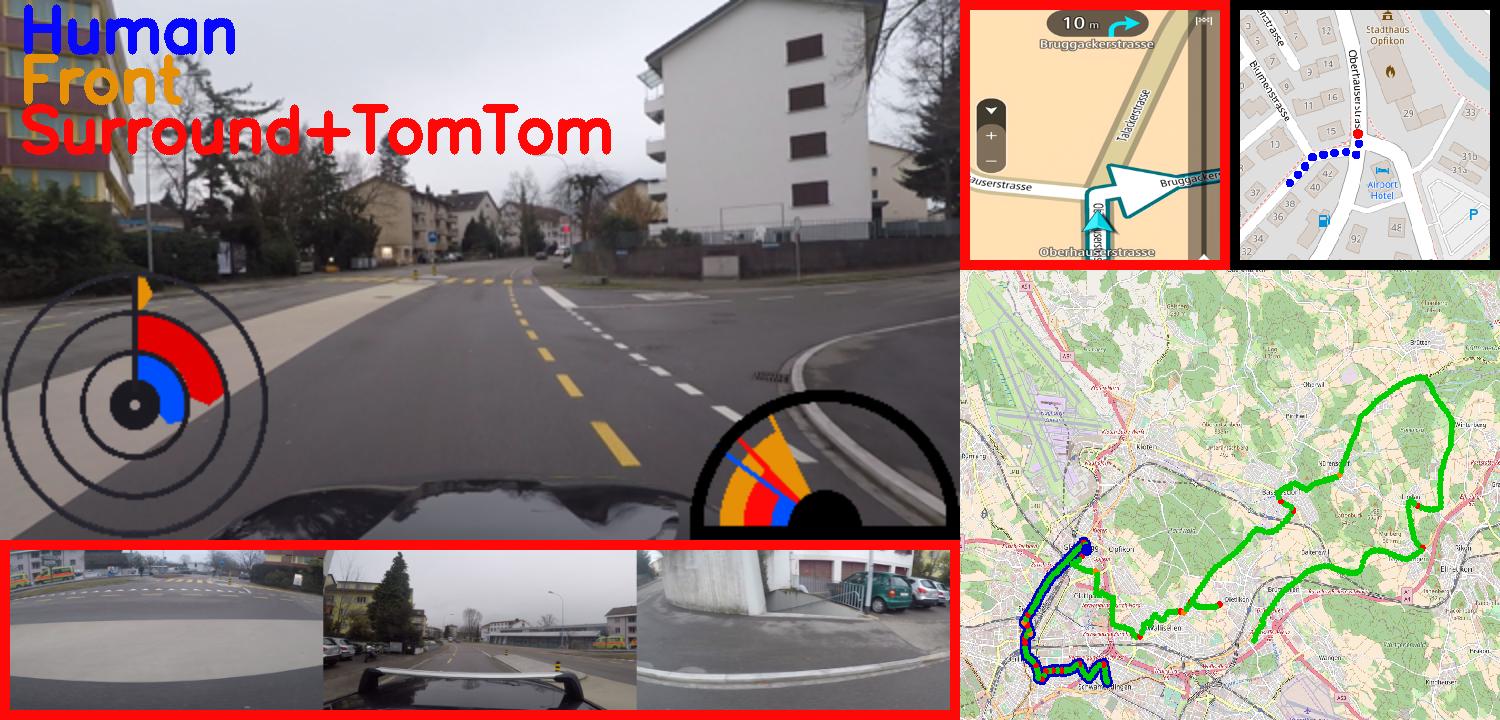}
\includegraphics[width=0.50\linewidth]{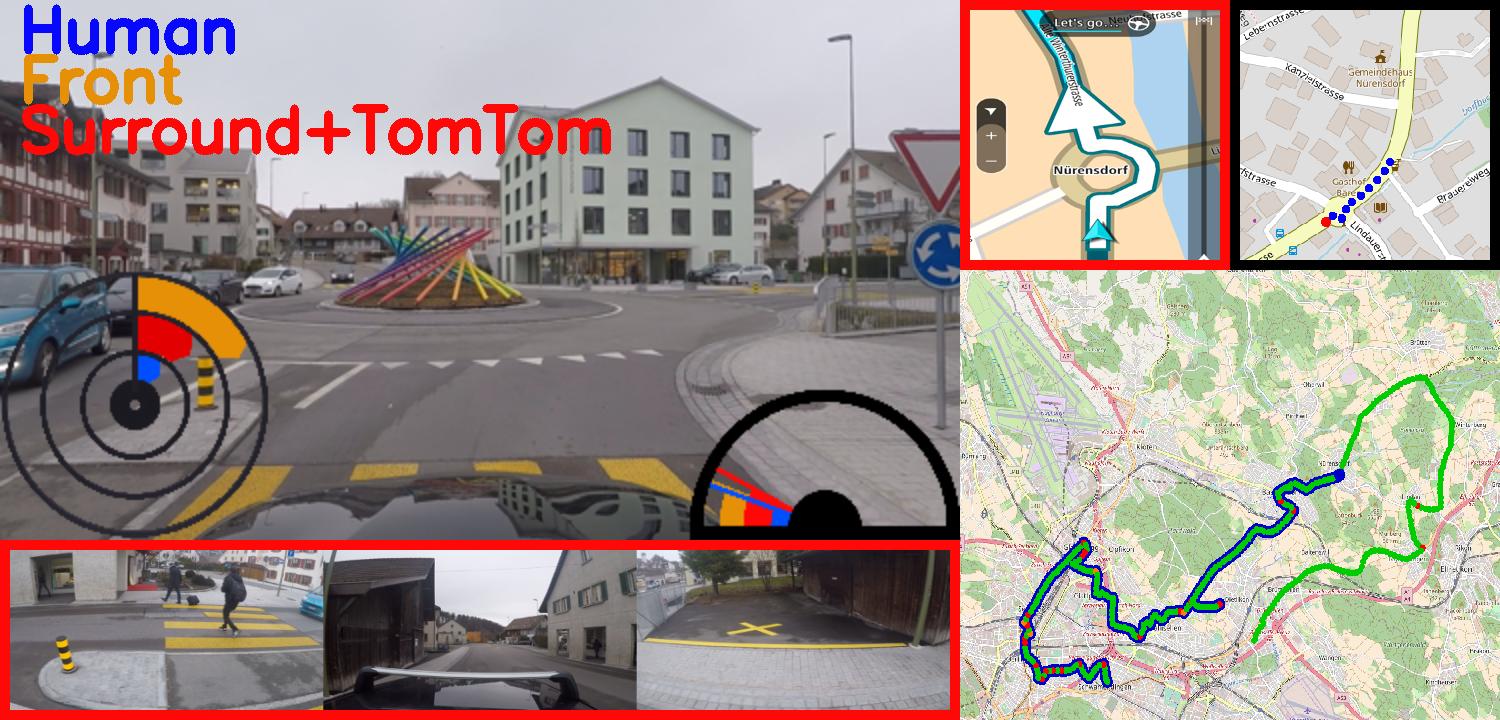} \\ 
\includegraphics[width=0.50\linewidth]{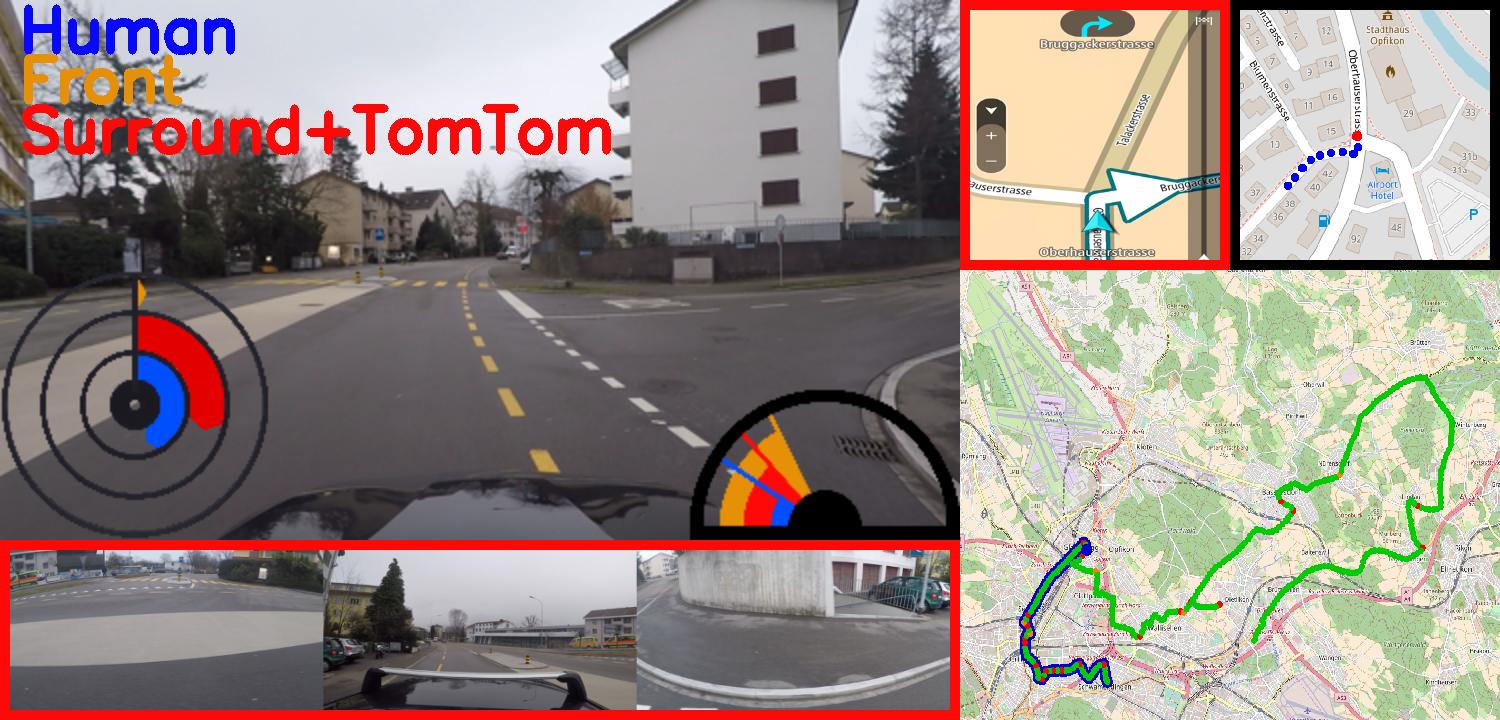}
\includegraphics[width=0.50\linewidth]{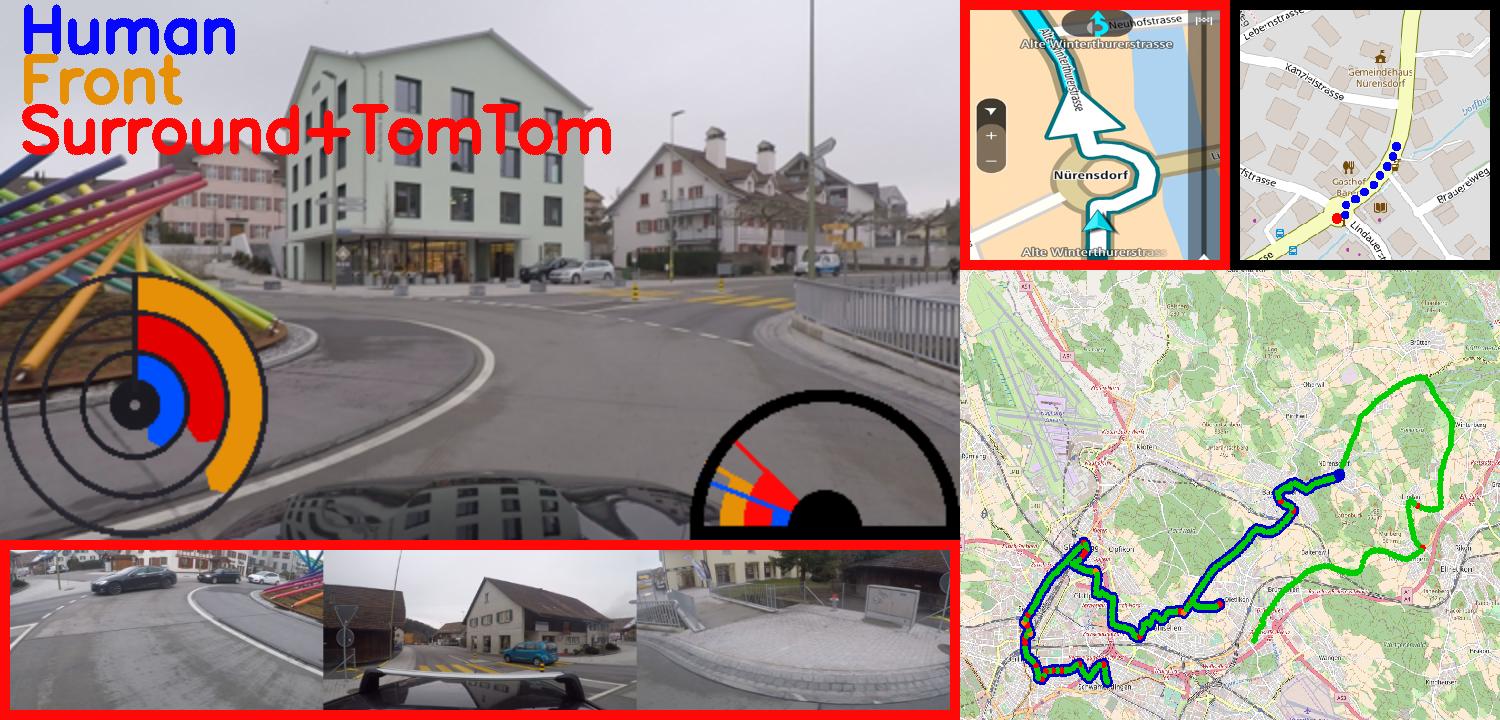} \\ 
\includegraphics[width=0.50\linewidth]{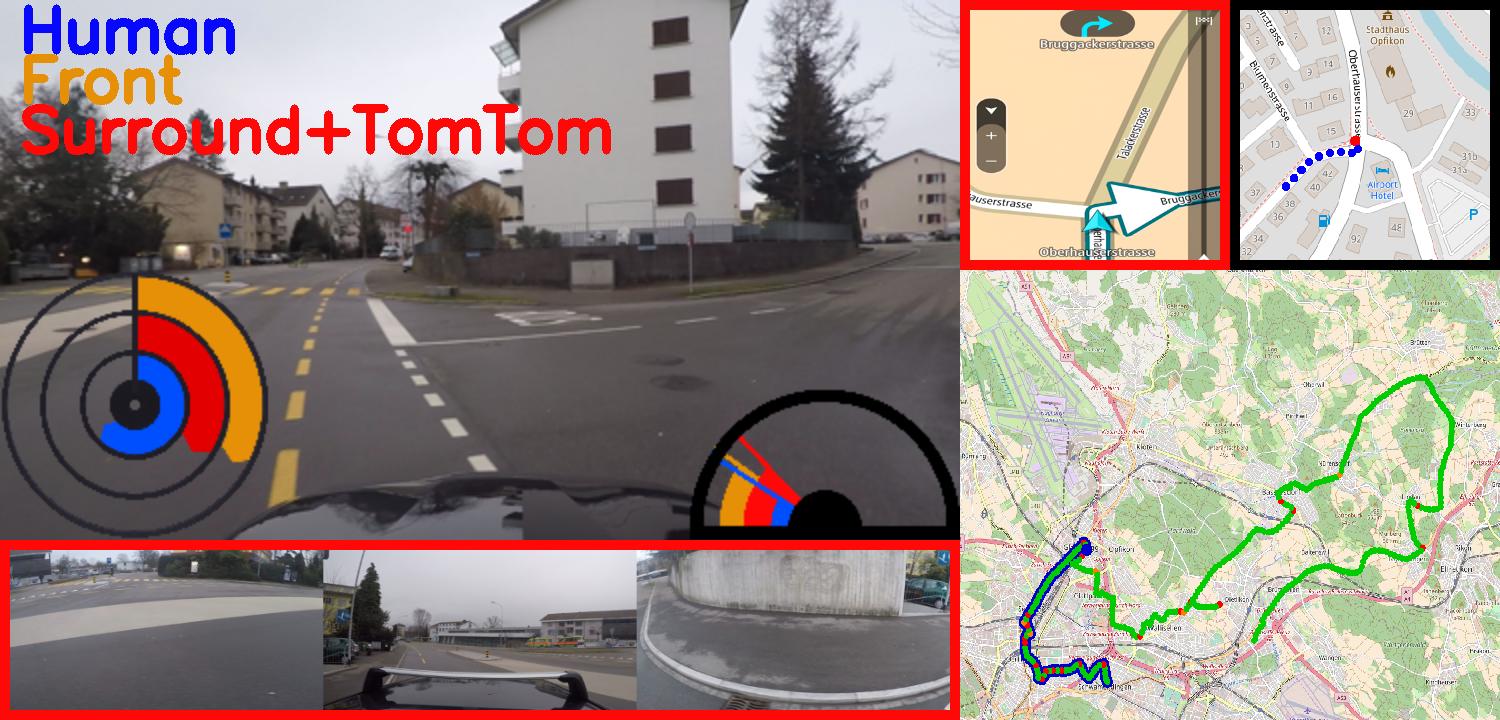}
\includegraphics[width=0.50\linewidth]{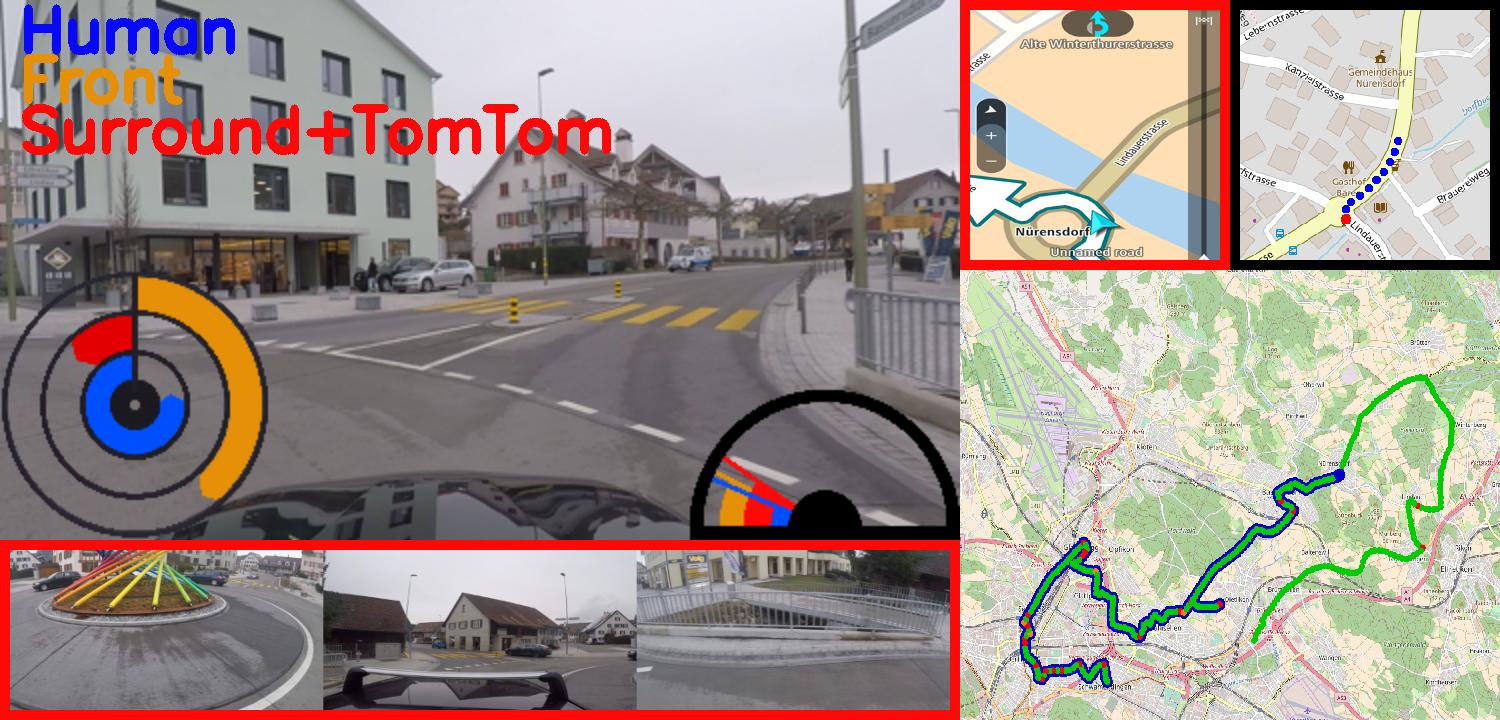} \\
\includegraphics[width=0.50\linewidth]{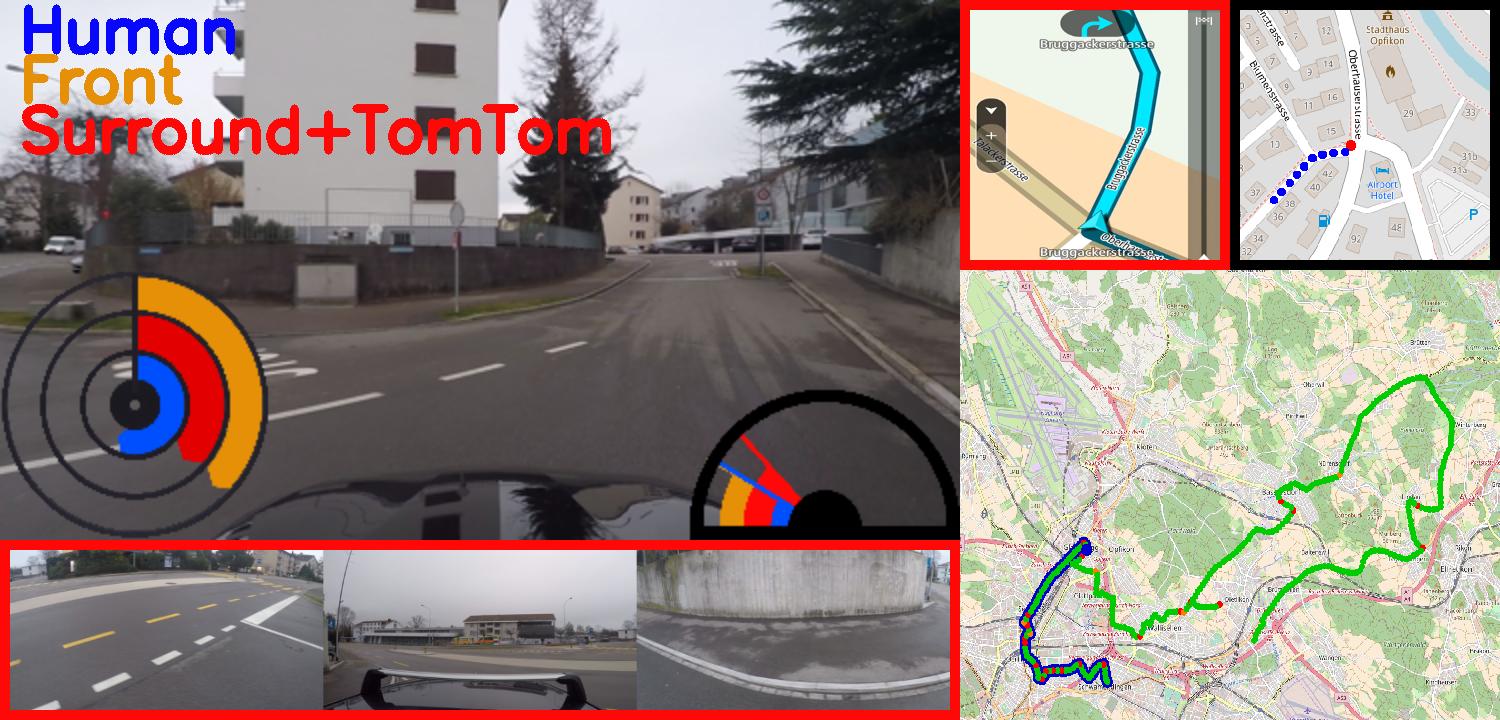}
\includegraphics[width=0.50\linewidth]{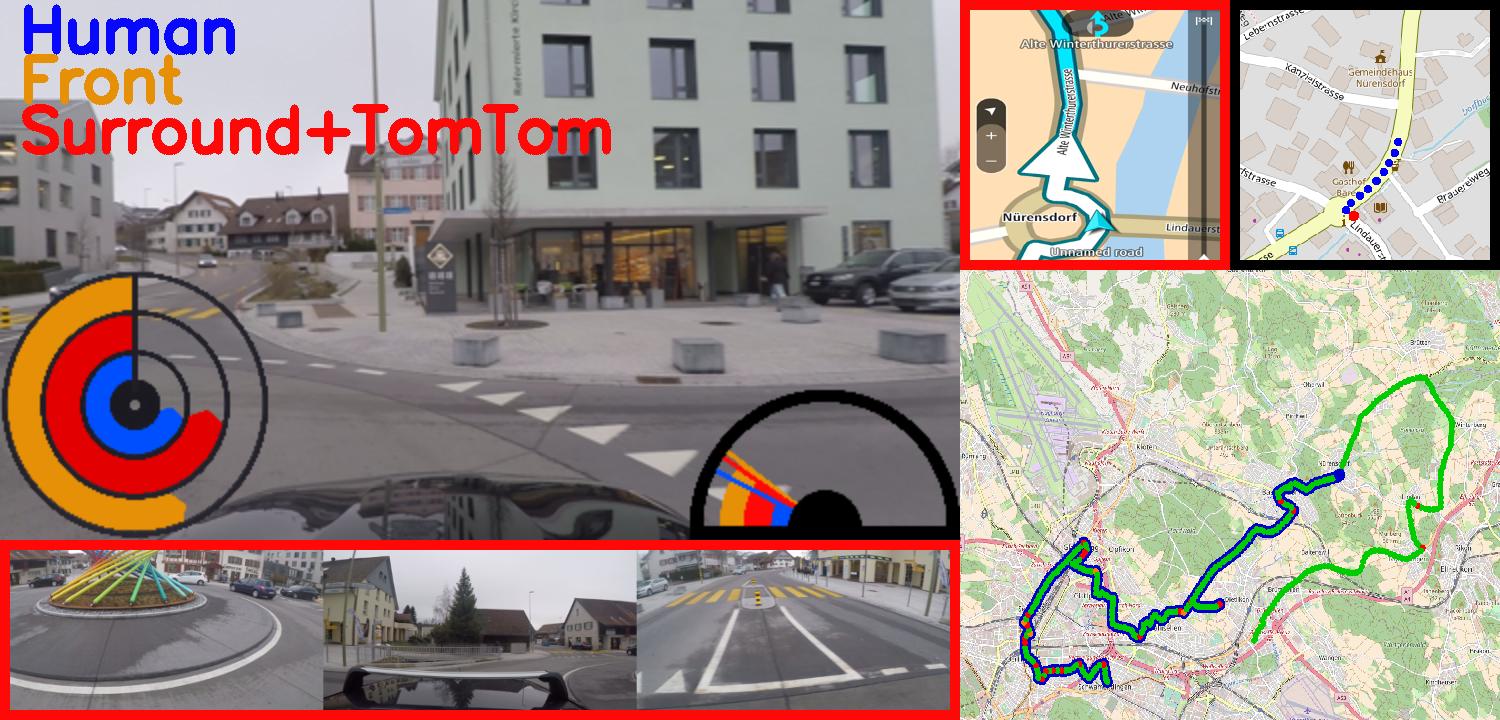} \\
\includegraphics[width=0.50\linewidth]{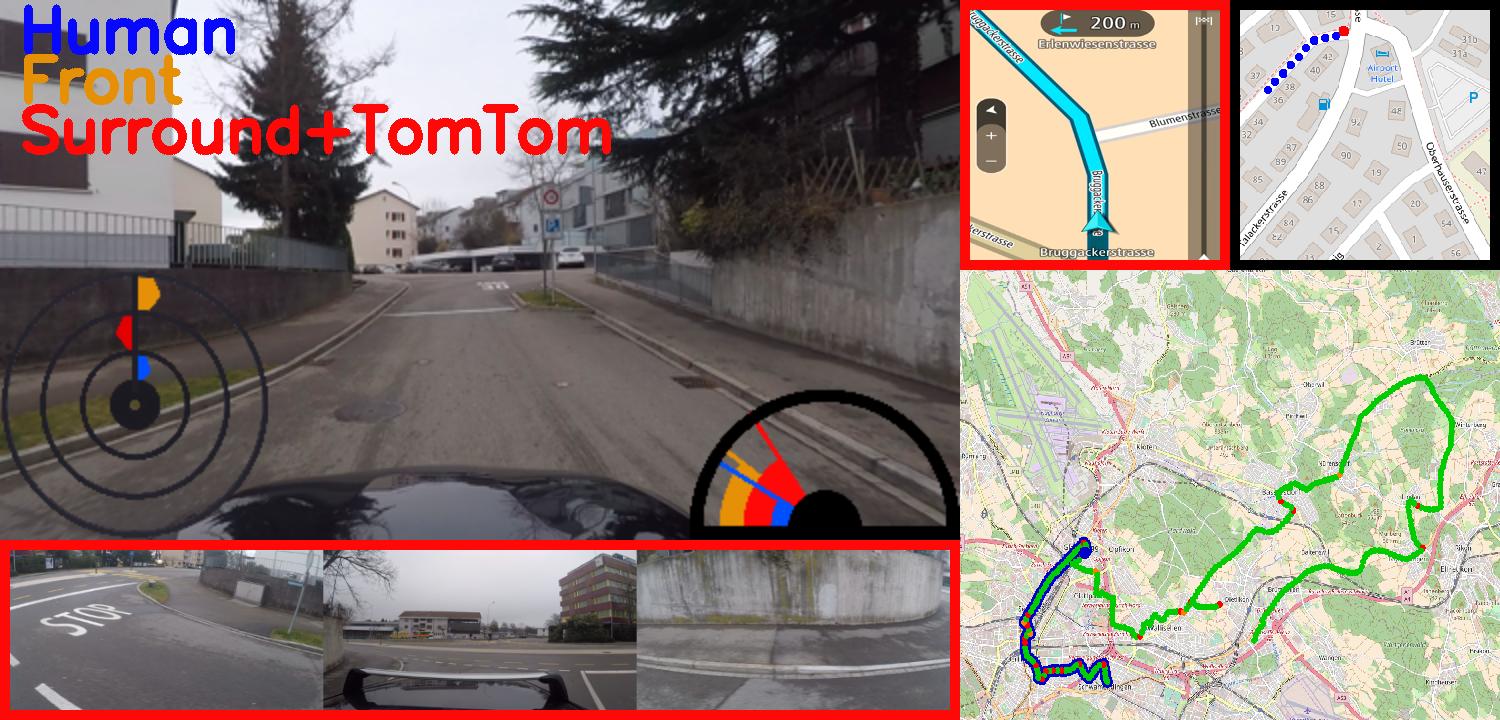}
\includegraphics[width=0.50\linewidth]{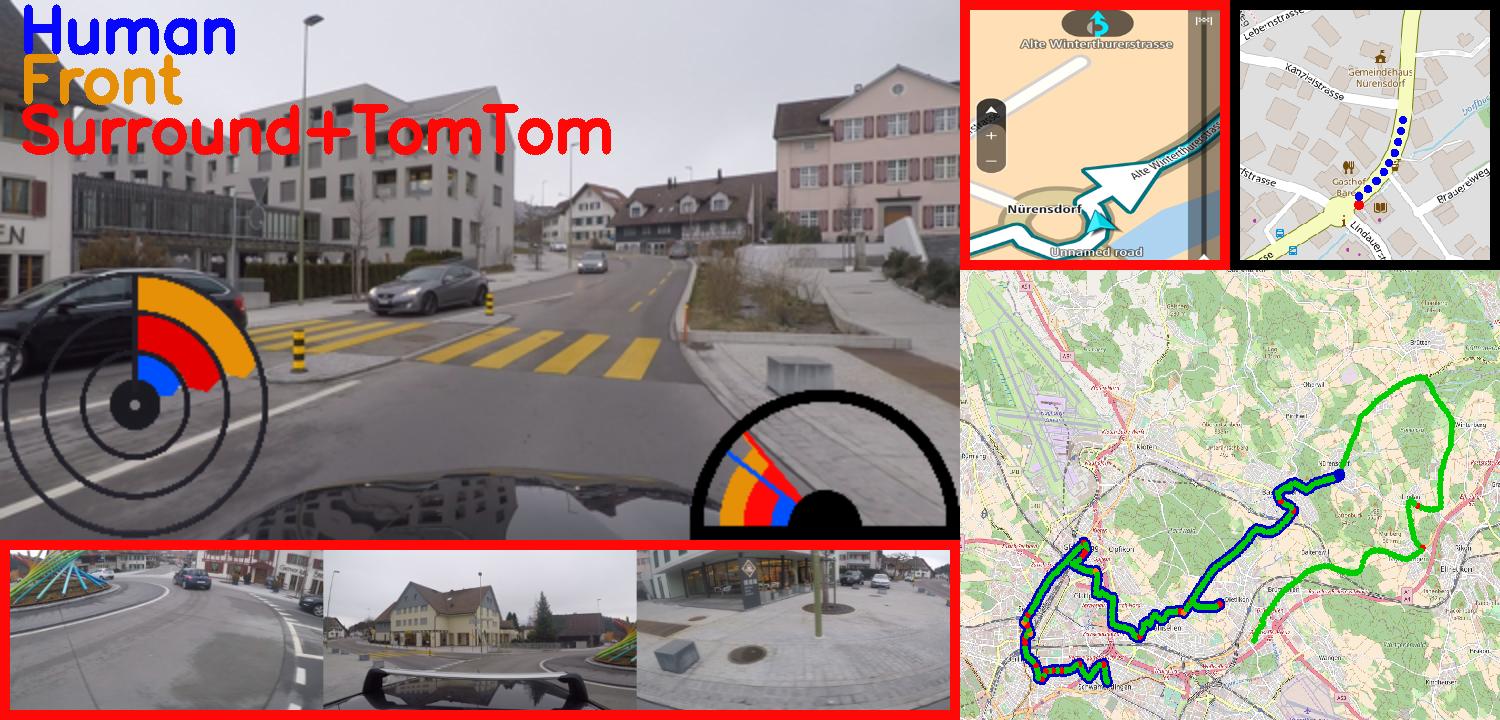} \\
\hspace{-75mm} \text{(1) right turn} & \hspace{-45mm} \text{(2) roundabout}   \\
\end{tabular}$
 \caption{Qualitative evaluation of Surround-View + TomTom and Front-Camera-Only models. Example for two driving maneuvers: (1) right turn (2) roundabout, with a sequence of temporal frames. Better seen on screen.} 
\label{fig:examples:pics} 
\end{figure*}

\clearpage

\end{document}